\documentclass{article}

\usepackage[preprint,nonatbib]{cml4impact_2022}

\usepackage{hyperref}
\usepackage{url}

\usepackage[utf8]{inputenc} % allow utf-8 input
\usepackage[T1]{fontenc}    % use 8-bit T1 fonts
\usepackage{hyperref}       % hyperlinks
\usepackage{url}            % simple URL typesetting
\usepackage{booktabs}       % professional-quality tables
\usepackage{amsfonts}       % blackboard math symbols
\usepackage{nicefrac}       % compact symbols for 1/2, etc.
\usepackage{microtype}      % microtypography
\usepackage{xcolor}         % colors

\usepackage{url}            % simple URL typesetting
\usepackage{nicefrac}       % compact symbols for 1/2, etc.

\usepackage{latexsym,amssymb,amstext,amsfonts,amsmath,graphicx,setspace,mathtools,bm,color}
\usepackage{cleveref}
\usepackage{caption,subcaption}
\usepackage{comment}
\usepackage{gensymb}

\usepackage{enumerate} 
\usepackage{paralist}
\usepackage{float}
\usepackage{array}
\usepackage{adjustbox}
\usepackage{boldline}

\usepackage{multirow}

\usepackage{lipsum}
\usepackage{abdalmageed_style}
\usepackage{enumitem}
\usepackage{wrapfig}
\setlist{topsep=0pt,parsep=0pt,partopsep=0pt}
\newcommand{\etal}[1]{\textit{et al.}~(\citeyear{#1})}

\newcommand{\rewrite}[1]{{\color{black}{#1}}}

\newcommand{\cdl}{CDL}

\title{Do-Operation Guided Causal Representation Learning with Reduced Supervision Strength}

\author{
Jiageng Zhu$^{1,2,3}$ \qquad Hanchen Xie$^{2,3}$ \qquad Wael AbdAlmageed$^{1,2,3}$\\
$^1$ USC Ming Hsieh Department of Electrical and Computer Engineering \\
$^2$ USC Information Sciences Institute \\
$^3$ Visual Intelligence and Multimedia Analytics Laboratory \\
\tt\small  \{jiagengz, hanchenx, wamageed\}@isi.edu}

\def\etal{\emph{et al.}}

\def\cg{causal graph}
\def\crelation{causal relationship}

\begin{document}
% \author{Jiageng Zhu\inst{1,2,3} \and Hanchen Xie\inst{2,3} \and Wael Abd-Almageed\inst{1,2,3}}
% \authorrunning{J. Zhu, H. Xie and W. Abd-Almageed}

% \institute{USC Ming Hsieh Department of Electrical and Computer Engineering \and
% USC Information Sciences Institute \and Visual Intelligence and Multimedia Analytics Laboratory\\
% \email{\{jiagengz, hanchenx, wamageed\}@isi.edu}}

\maketitle

\begin{abstract}
Causal representation learning has been proposed to encode relationships between factors presented in the high dimensional data. However, existing methods suffer from merely using a large amount of labeled data and ignore the fact that samples generated by the same causal mechanism follow the same causal relationships. In this paper, we seek to explore such information by leveraging \emph{do-operation} for reducing supervision strength. We propose a framework which implements \emph{do-operation} by swapping latent cause and effect factors encoded from a pair of inputs. Moreover, we also identify the inadequacy of existing causal representation metrics empirically and theoretically and introduce new metrics for better evaluation. Experiments conducted on both synthetic and real datasets demonstrate the superiorities of our method compared with state-of-the-art methods.
% Causal representation learning has been proposed to encode causal relationships between factors presented in the high dimensional data. Existing methods are limited to being trained and fully supervised by ground-truth generative factors. In this paper, we seek to reduce supervision strength by leveraging intervention on either the cause factor or effect factor for reducing supervision strength. Applying interventions on cause factors and effect factors will lead to different results since intervention on effect factors will change the causal graph. In contrast, intervention on cause factors will not change the relationships. The intervention can also be called  \emph{do-operation}. Based on this attribute of \emph{do-operation}, we propose a framework called Do-VAE, which implements \emph{do-operation} by swapping latent cause factors and effect factors encoded from a pair of inputs and utilizing the supervision signal from a pair of inputs by comparing original inputs and reconstructions. Moreover, we also identify the inadequacy of existing causal representation metrics and introduce new metrics for better evaluation.
\end{abstract}

\section{Introduction}
\label{sec:introduction}

\begin{wrapfigure}{r}{0.4\textwidth}
\vspace{-\baselineskip}
\centering
%\fbox{\rule{0pt}{2in}
%\rule{0.9\linewidth}{0pt}}
   \includegraphics[width=0.35\textwidth]{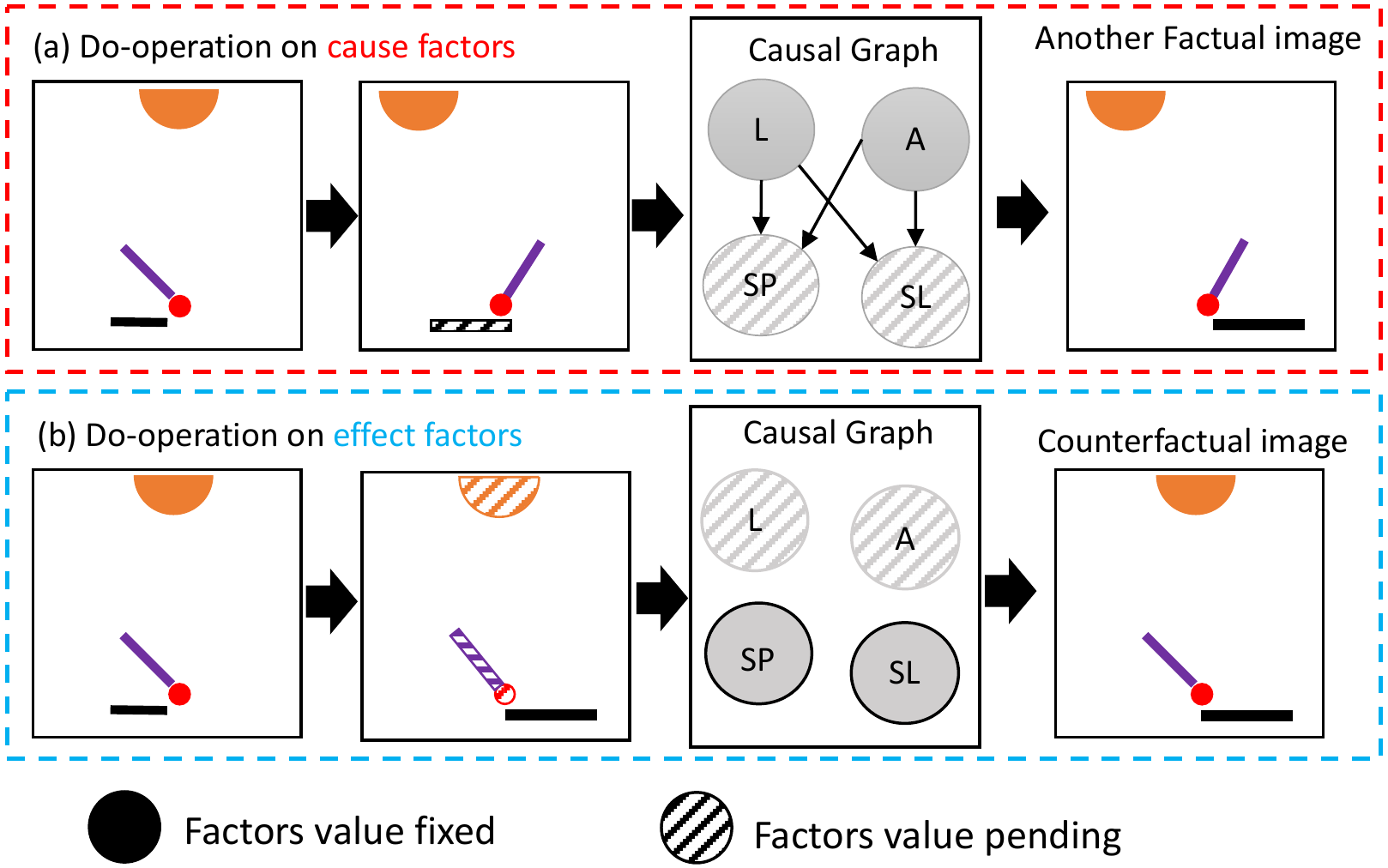}
   %\vspace{-3pt}
   \caption{\rewrite{ \emph{Do-operation} to cause and effect factors. Light position (L) and pendulum angle (A) are the cause of shadow position (SP) and shadow length (SL).  Applying \emph{do-operation} to cause factors will change the effect factors accordingly. Oppositely, applying \emph{do-operation} to the effect factors will not affect the cause factors, and the original causal relationships from L and A to SP and SL are removed. Thus, a counterfactual sample will be created.}}
\label{fig:introduction}
\vspace{-\baselineskip}
% \vspace{-.6cm}
\end{wrapfigure}
Causal representation learning \cite{bib:towards-causal} has been proposed to extract causal relations from high dimension observations. To this end, CausalVAE \cite{DBLP:conf/cvpr/YangLCSHW21} contains a causal layer and a mask layer as parts of deep neural network (DNN) architecture, and uses labels of generative factors to learn the causal relationship between different latent factors.

However, training CausalVAE requires labels of all generative factors, which may still pose a strong assumption. For instance, all semantic causal factors need to be carefully annotated, which is either costly or hard to be identified in the first place. Further, since it relies on full supervision, CausalVAE limits the dimensionality of the latent representation to be the same as the number of generative factors and leaves no space for other \emph{unknown} confounding factors which can be entangled with semantically meaningful latent factors and harm the performance. \rewrite{Moreover, CausalVAE incorporates ground-truth generative factors so that causal layer of CausalVAE can be trained separately from VAE. Thus, rather than extracting causal relations from high dimension observations, causal representation in CausalVAE is merely obtained through ground-truth generative factors, which is also used as a part of inputs during training.}
% The main advantage of CausalVAE, compared with previous work, is that it considers causal relationships between the semantic factors in the latent representations. 
% Previous models, such as $\beta$-VAE \cite{DBLP:conf/iclr/HigginsMPBGBML17}, LadderVAE \cite{bib:laddervae} and ConditionVAE \cite{bib:conditionvae} also encode semantic information in the latent representations but simply assume that the semantic factors are independent of each other.

To avoid the constraint of using fully supervised training, we utilize the \emph{do-operation}, illustrated in \Cref{fig:introduction}, to learn causal representation with reduced supervision. \emph{Do-operation} \cite{reason:Pearl09a} defines an intervention that remove certain relationships in the causal graph and replace a factor with a constant. According to Pearl \etal~\cite{reason:Pearl09a}, the causal effects can only propagate from cause factors to effect factors and not inversely. Thus, when \emph{do-operation} is applied to cause factors, a new and factual sample will be generated. Conversely, when \emph{do-operation} is applied to effect factors, the cause factors should be unaffected. Further, since \emph{do-operation} changes the values of effect factors to constants, the newly generated sample can be counterfactual. 
% We hypothesize that utilizing the \emph{do-operation} during model training and separately applying \emph{do-operation} to cause and effect factors can encourage the model to learn the correct causal relationship and prevent the model from encoding false \crelation. 
When training the model, since the supervision strength is reduced to limited or even no labels, we use two latent representations encoded from a pair of inputs and apply \emph{do-operation} via exchanging their latent factors with each other. By comparing the new reconstructions after \emph{do-operation} with the original inputs, a supervision signal can be introduced. 
% to learn the correct causal representation.

% \begin{figure}[]
%     \centering
%     \includegraphics[width=1\textwidth]{figures/introduction.pdf}
%     \caption{\rewrite{ \emph{Do-operation} to cause and effect factors. Light position (L) and pendulum angle (A) are the cause of shadow position (SP) and shadow length (SL).  Applying \emph{do-operation} to cause factors will change the effect factors accordingly and another factual sample will be produced. Oppositely, applying \emph{do-operation} to the effect factors will not affect the cause factors. Further, since the effect factors are replaced by constants, the original causal relationships from L and A to SP and SL are removed. Thus, a counterfactual sample will be created.}}
%     \label{fig:introduction}
% \end{figure}
CausalVAE \cite{DBLP:conf/cvpr/YangLCSHW21} uses MIC and TIC \cite{bib:mic-tic} to evaluate the performance of causal representation learning. However, MIC and TIC only calculate mutual information between the latent representation and its corresponding ground truth generative factors. We argue, therefore, that MIC and TIC can only reflect the correctness of the marginal distribution of each factor itself, whereas no causal relationship between factors can be measured. Therefore, we propose new metrics for better evaluation.
% to better evaluate causal representation learning models.

% To address the limitations of both CausalVAE and current evaluation metrics, we propose a new framework which reduces the strength of supervision signal and learns to encode confounding factors present in the input data, as well as new evaluation metrics that can better evaluate the performance of causal representation learning. Our experiments show that no label is required on synthetic datasets and only a small amount of labels are needed for real face images. 

% The main contributions of our work are:
% \begin{itemize}
%     \item A novel model architecture incorporating causal discovery layer which discovers causal relationships of latent causal factors.
%     \item A training algorithm which reduces the strength of the supervision signal through applying \emph{do-operation} to latent cause and effect factors encoded from pair of inputs.
%     \item Identifying the weaknesses of existing causal representation learning  metrics  and introducing new metrics for better evaluation.  
%     \item Comprehensive experiments are conducted on both synthetic and real datasets, where the results empirically demonstrate the superiorities of our method.
% \end{itemize}

% \section{Related Work}
\label{sec:related_work}

\begin{figure}[]
    \centering
    \includegraphics[width=0.8\textwidth]{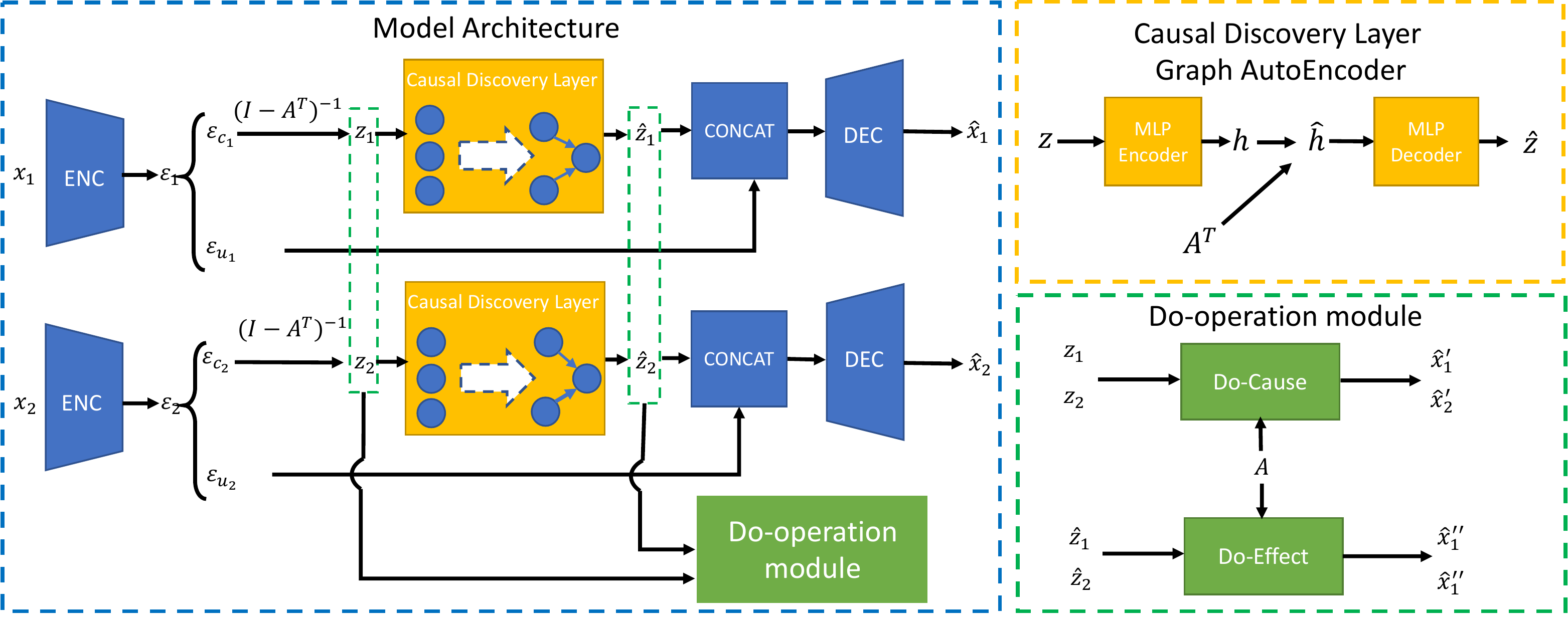}
    \caption{Model structure. The input $x$ is encoded to exogenous variable $\varepsilon$, which can be  further splits into latent causal factors $\varepsilon_c$ and unknown nuisance factors $\varepsilon_u$. $\varepsilon_c$ is then mapped to endogenous variable $z$. The causal relationships are discovered and calculated through causal discovery layer. The unknown nuisance factors $\varepsilon_u$ and causal representation $\hat z$ is then concatenated as the inputs of a decoder. A pair of inputs are used to introduce supervision signal. Two encoders and two decoders in model share same weights respectively.  }
    \label{fig:model-architecture}
\end{figure}
\textbf{Related work:} Disentangled representation learning aims at attaining mutual independent latent factors \cite{bib:bengio_representaion} and Variational Autoencoder (VAE) \cite{DBLP:journals/corr/KingmaW13} is the basic framework of most disentanglement methods, where the loss function is $L_{VAE}(x,z)  = -\mathbb{E}_{q_{\phi}(z|x)}[logp_{\theta}(x|z)]  +  D_{KL}(q_{\phi}(z|x)||p(z))$. Other unsupervised VAEs including $\beta$-VAE \cite{DBLP:conf/iclr/HigginsMPBGBML17} AnnealedVAE \cite{burgess2018understanding}, LadderVAE \cite{bib:laddervae} and $\beta$-TCVAE \cite{chen2019isolating} are proposed by modifying $L_{VAE}$. Causal representation learning is the extension of disentangled representation learning. To achieve causal representation, CausalVAE \cite{DBLP:conf/cvpr/YangLCSHW21}, built upon iVAE \cite{bib:ivae}, removes the requirement of prior knowledge of true causal graph by introducing the causal layer and mask layer into the model. However, all generative factor labels are required to train CausalVAE.
% Brehmer \etal~\rewrite{\cite{bib:weakly-causal} proposes LCM model to learn causal representation under weakly supervised training. To train LCM, a dataset is required to contain both factual data and counter-factual data.}

\section{Method}
\label{sec:method}

\textbf{Model architecture:} We propose a new architecture, shown in \Cref{fig:model-architecture}, as well as a training algorithm that greatly reduces the supervision strength via a \emph{do-operation} module. We use $x$ denotes an input image, $\varepsilon = [\varepsilon_c, \varepsilon_u ]$ denotes exogenous latent factors which is further split into causal and unknown nuisance exogenous factors, $z$ denotes latent causal factors, $\hat z$ denotes latent causal factors after causal discovery and $\hat x$ denotes reconstructed images.

In contrast with CausalVAE, our framework uses $\varepsilon_u$ to encode unknown  nuisance factors. Meanwhile, similar to CausalVAE, the exogenous factors $\varepsilon_c$  are first transformed to endogenous latent factors $z$, and a causal discovery layer (\cdl) propagates causal effects from parent factors to their child factors. We use a graph autoencoder (GAE) \cite{bib:gae} as \cdl, which learns nonlinear causal relationships and thus generalizes over NOTEARS \cite{bib:notears} used in CausalVAE. The unknown nuisance latent factors $\varepsilon_u$ are concatenated with the latent causal factors $\hat z$ as the input of a decoder. As discussed in \cite{locatello2019challenging}, unsupervised learning can not identify expected latent representations so that supervision is necessary. To reduce supervision strength in CausalVAE and inspired by \cite{Locatello2020Disentangling}, we use a pair of inputs and implement \emph{do-operation} during training to utilize a weak supervision signal. The \cdl{}, that applies causal effect from parent factors to child factors, is the key to implement \emph{do-operation}, descried in \Cref{sec:do-opeartion}, in order to decrease supervision strength. By using this new training strategy, we show in \Cref{sec:evaluation} that no label is needed during training on synthetic datasets and only a small amount of labels is needed on real datasets.

\begin{wrapfigure}{l}{0.38\textwidth}
\vspace{-\baselineskip}
\centering
%\fbox{\rule{0pt}{2in}
%\rule{0.9\linewidth}{0pt}}
   \includegraphics[width=0.35\textwidth]{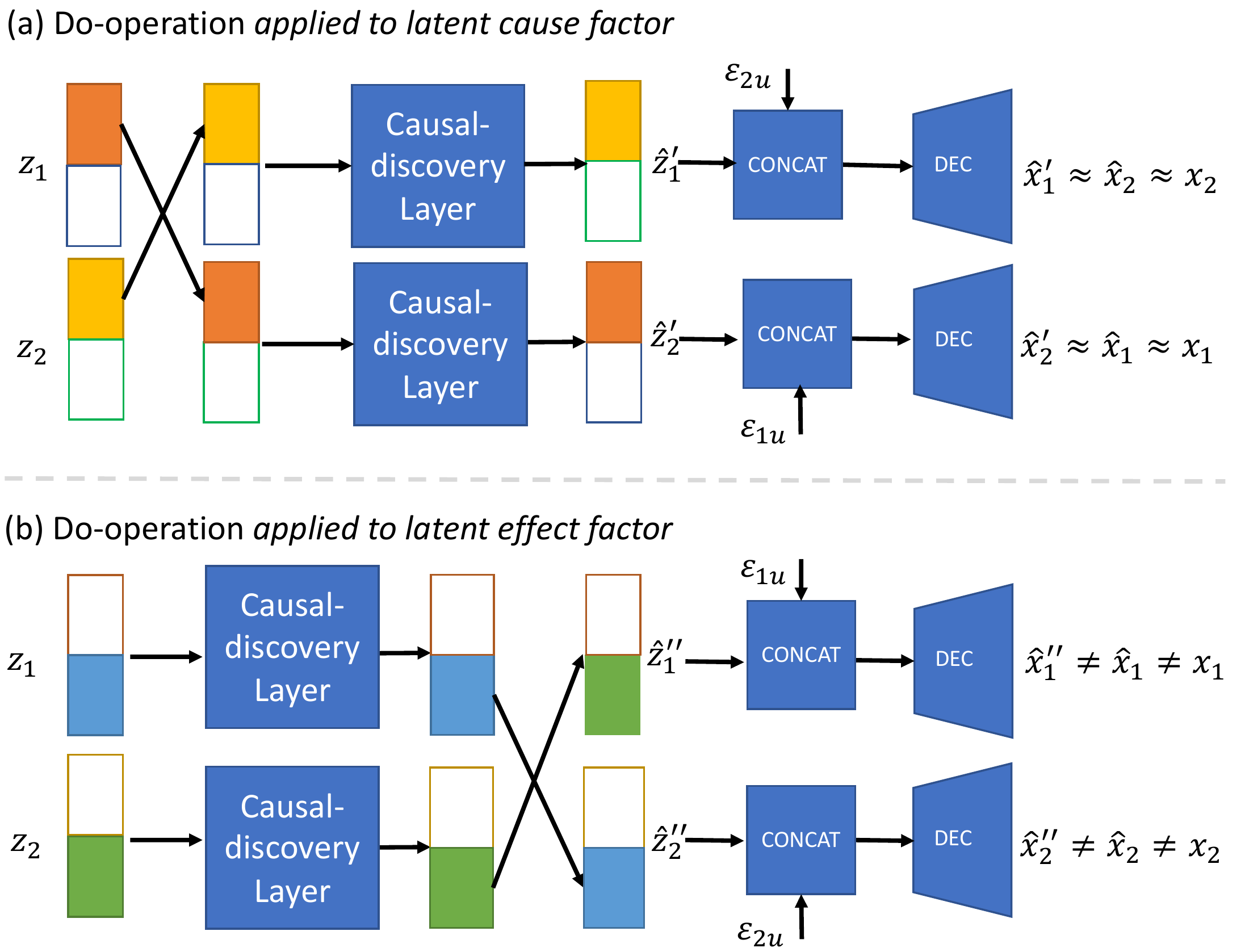}
   \vspace{-5pt}
   \caption{\emph{Do-operation} on cause factors encourage model to learn correct causal relationships, while \emph{do-operation} on effect factors prevent model learning wrong causal relationship.}
\label{fig:do-operation}
% \vspace{-.6cm}
\vspace{-\baselineskip}
\end{wrapfigure}
\leavevmode

% \begin{figure}[]
% \centering
% \begin{subfigure}[h]{0.45\textwidth}
%     \centering
%     \includegraphics[width=0.8\textwidth]{figures/cause_factors.pdf}
%     \caption{\emph{Do-operation} applied to cause factors.}
%     \label{fig:do-on-cause}
% \end{subfigure}
% \hfill
% \begin{subfigure}[h]{0.45\textwidth}
%     \centering
%     \includegraphics[width=0.8\textwidth]{figures/effect_factors.pdf}
%     \caption{\emph{Do-operation} applied to effect factors.}
%     \label{fig:do-on-effect}
% \end{subfigure}
% \caption{ \emph{Do-operation} on cause factors encourage model to learn correct causal relationships, while \emph{do-operation} on effect factors prevent model learning wrong causal relationship. }
% \label{fig:do-operation}
% \end{figure}
% \subsection{Do-operation During Training}

\label{sec:do-opeartion}
\textbf{Do-operation illustrates  \crelation:} \emph{Do-operation} \cite{reason:Pearl09a} defines an intervention that deletes a specific relationship in the causal graph and replaces factors with constants. 
As shown in \Cref{fig:introduction}, if \emph{do-operation} is applied to cause factors, since the original causal graph stays unchanged, the effect factors will be affected according to their parent factors. Conversely, when \emph{do-operation} is applied to effect factors, cause factors will not affect the value of effect factors. This process can be shown in \Cref{eq:do-operation1,eq:do-operation2}.
\begin{align}
    &do(z_c^{(l)})  := z_c^{(m)};  ~ ~~  f([do(z_c^{(l)}), z_e^{(l)}]) = [z_c^{(m)}, z_e^{(m)}]; \hfill \label{eq:do-operation1} \\
    &do(z_e^{(l)})  := z_e^{(m)};  ~  ~~ f([z_c^{(l)}, do(z_e^{(l)})]) = [z_c^{(l)}, ~~z_e^{(m)}]; \label{eq:do-operation2}
\end{align}
where $f$ is the causal relationship function, and generative factors $z$ are split into cause factors $z_c$ and effect factors $z_e$. By assigning previous cause factors $z_c^{(l)}$ with new value $z_c^{(m)}$, effect factors $z_e$ will change accordingly. Oppositely, if \emph{do-operation} is applied to effect factors $z_e^{(l)}$  whose value is replaced  by $z_e^{(m)}$, cause factors $z_c$ should stay unchanged. Besides, The output of causal function $f$ can be counterfactual since the original causal relationship has changed.
% \rewrite{However, during training, the ground-truth causal relation functions are unknown, and the latent factors are not guaranteed to contain the expected semantic meaning. Thus,  directly utilizing \Cref{eq:do-operation1,eq:do-operation2} to discover causal relationships is unfeasible. To solve these issues, we need to incorporate reconstruction as down-stream task. During training, whether one specific factor is cause or effect is determined by the learnable causal matrix $A$ in the causal discovery layer. Besides, as mentioned in \Cref{sec:introduction}, a pair of inputs are used to introduce the supervision signal. To utilize this supervision signal during training and encourage meaningful semantic information is encoded into latent space when applying \emph{do-operation} to a factor, the new value set to that factor is chosen from the corresponding latent factor encoded from another input in a pair. By comparing the reconstructions with the original inputs, the model is encouraged to learn both semantic information and causal relations between each semantic factor. Details of this training process are described in the following paragraphs.}

\textbf{Do-operation on cause factors (Do-Cause):}
As shown in the \Cref{eq:do-operation1}, if we apply \emph{do-operation} to cause factors $z_c$, since $z_c$ have no parent factors, the causal graph is unchanged and the value of effect factors $z_e$ should change accordingly. To train our model, since no label or limited labels of generative factors are available, we use  pairs of images as a weak supervision signal to encourage the model to learn causal representation.  As illustrated in \Cref{fig:do-operation}a, except the regular propagation of inputs, after two endogenous latent factors $z_1$ and $z_2$ are encoded from a pair of inputs $x_1$ and $x_2$,  we exchange the cause factors of two latent representations with each other to create two new latent representation $z_1^{\prime} =[do(z_{1c}), z_{1e} ] $ and $z_2^{\prime} =[do(z_{2c}), z_{2e}] $, where a latent factor is cause or effect is determined by the learnable causal matrix $A$ in CDL.  As shown in \Cref{eq:cause-exchange}, two new representations $z_1^{\prime}$ 
and $z_2^{\prime}$ are fed into the \cdl{} and then concatenated with their corresponding unknown nuisance factors $\varepsilon_{u_{1}}$ and $\varepsilon_{u_{2}}$ as inputs of the decoder.
\begin{equation}
\begin{split}
    \label{eq:cause-exchange}
    z_1^{\prime} := [do(z_{c_1}), z_{e_1}] & = [z_{c_2}, z_{e_1} ]; ~~~ \hat z_1^{\prime} = f(z_1^{\prime}); ~  ~~ \hat x_1^{\prime} = Dec(\hat z_1^{\prime}, \varepsilon_{u_2} ) \\
    z_2^{\prime} := [do(z_{c_2}), z_{e_2}] & = [z_{c_1}, z_{e_2}]; ~~~ \hat z_2^{\prime} = f(z_2^{\prime}); ~  ~~ \hat x_2^{\prime}  =  Dec( \hat z_2^{\prime} , \varepsilon_{u_1} ) \\ 
\end{split}
\end{equation}
Recall that from \Cref{eq:do-operation1}, the new outputs of \cdl{} should be same with the original outputs of \cdl, where $\hat z_1^{\prime} = \hat z_2$ and $\hat z_2^{\prime} = \hat z_1$, since \emph{do-operation} on cause factors does not change \cg, and the unchanged causal graph propagates causal relationships from cause factors to effect factors. Since the new latent causal representation $\hat z_2^{\prime}$ and $\hat z_1^{\prime}$ should be same with the original latent causal representation $z_1$ and $z_2$, their corresponding reconstructions $\hat x_1^{\prime}$ and $\hat x_2^{\prime}$ after the decoder should also be same with the the original inputs $x_2$, $x_1$. As shown in \Cref{eq:cause-loss}, by comparing new reconstructions with the original inputs, the model is encouraged to learn the correct causal relationships, where $d$ is distance function, such as binary cross entropy or mean square error.
\begin{equation}
\label{eq:cause-loss}
L_{cause} = d(\hat x_1^{\prime}, x_2) + d(\hat x_2^{\prime}, x_1)
\end{equation}
\textbf{Do-operation on effect factors (Do-Effect):}
Compared with \emph{do-operation} on the cause factors, since the causal graph will change when applying \emph{do-operation} to the effect factors, the latent effect factors should be exchanged after the \cdl{} in order to remove the effect of cause factors. The whole process of \emph{do-operation} on the effect factors can be shown in \Cref{eq:effect-exchange}.
\begin{equation}
\begin{split}
    \label{eq:effect-exchange}
    \hat z_1 = f(z_1); ~~~ \hat z_1^{\prime\prime} := [\hat z_{c_1}, do(\hat z_{e_1})] & =[\hat z_{c_1}, \hat z_{e_2}];  ~~~ \hat x_1^{\prime\prime}   = Dec(\hat z_1^{\prime\prime}, \varepsilon_{u_1} ) \\
    \hat z_2 = f(z_2); ~~~ \hat z_2^{\prime\prime} := [\hat z_{c_2}, do(\hat z_{e_2})] & =[\hat z_{c_2}, \hat z_{e_1}];  ~  ~~ \hat x_2^{\prime\prime}  =  Dec(\hat z_2^{\prime\prime}, \varepsilon_{u_2} ) \\ 
\end{split}
\end{equation}
Since \emph{do-effect} changes the existing causal graph, the new latent representations $\hat z_1^{\prime\prime}$ and $\hat z_2^{\prime\prime}$ are not consistent with their corresponding  latent representations $z_1$ and $z_2$. Thus, after decoder, the new reconstructions $\hat x_1^{\prime\prime}$ and $\hat x_2^{\prime\prime}$ will be different from their original inputs $x_1$ and $x_2$. Further, the new reconstructions are are actually counterfactual images as illustrated in \Cref{fig:introduction}b. In practice, using MSE or BCE may lead to degenerated solution where $\hat x^{\prime\prime}$ are random noise. To solve this issue, we use a classifier $C_w$ to distinguish factual images, including $x_i$, $\hat x_i$ and $\hat x_i^{\prime}$, with counterfactual images $\hat x_i^{\prime\prime}$, where classifier and VAE are trained alternatively. The losses of training classifier and \emph{do-operation} on the effect factors are shown in \Cref{eq:classifier-loss,eq:effect-loss} respectively.
\begin{align}
    &L_{cla} = \mathbf{BCE}(C_{w}(x_i),\mathbf{1}) + \mathbf{BCE}(C_{w}(\hat x_i),\mathbf{1}) + \mathbf{BCE}(C_{w}(\hat x_i^{\prime}),\mathbf{1}) + \mathbf{BCE}(C_{w}(\hat x_i^{\prime\prime}),\mathbf{0}) \hfill \label{eq:classifier-loss} \\
    &L_{effect} = \mathbf{BCE}(C_{w}(\hat x_1^{\prime\prime}),\mathbf{0}) + \mathbf{BCE}(C_{w}(\hat x_2^{\prime\prime}),\mathbf{0}) \label{eq:effect-loss}
\end{align}
\textbf{Training model with reduced supervision strength:}
As discussed in \Cref{sec:introduction} and empirically proven in \Cref{sec:evaluation}, our method only requires a small amount of supervision to train. For synthetic datasets, where actually no label are needed, the loss function is shown in \Cref{eq:no-label-loss}, where $L_{VAE}$ is same with $L_{VAE} $ shown in \Cref{sec:related_work}.
\begin{equation}
\label{eq:no-label-loss}
    L_{no-label} =  L_{VAE}(x, z) + \alpha L_{cause} + \beta L_{effect} + \gamma ||\hat z - z||^2_2 + h(A)
\end{equation}
where $\alpha$, $\beta$ and $\gamma$ are hyperparameters for regularizations. $||\hat z_i - z_i||^2_2$ is added to the  loss since the outputs of \cdl{} should align with their inputs.  $h(A)$ is an acyclicity constraint for the causal graph $A$. In our implementation, we use  $h(A) = tr(e^{A\odot A})-d$ as proposed in \cite{bib:notears}. 

% \notate{(I agree to remove label setting of real world datasets paragraph in experiments but I suggest keeping the description about the label loss in this section)}
If some labels of generative factors are available, similar to CausalVAE \cite{DBLP:conf/cvpr/YangLCSHW21}, we \rewrite{utilize} them by adding label constrains to \Cref{eq:no-label-loss} which leads to  \Cref{eq:semi-loss}, where $f$ is CDL.
\begin{equation}
\label{eq:semi-loss}
    L_{semi} = L_{no-label} + || u - f(u) ||_2^{2} + D_{KL}(q_{\phi}(z|x,u)||p(z|u))
\end{equation}

\section{Evaluation Metrics For Causal Representation Learning}
\label{sec:metric}

Maximum Information Coefficient (MIC) and Total Information Coefficient (TIC) \cite{bib:mic-tic} have originally been proposed as general purpose metrics to measure correlation between two random variables. Both metrics range from $0$ to $1$ and the higher value indicates better performance. 
\begin{wrapfigure}{l}{0.4\textwidth}
\vspace{-\baselineskip}
\centering
%\fbox{\rule{0pt}{2in}
%\rule{0.9\linewidth}{0pt}}
   \includegraphics[width=0.4\textwidth]{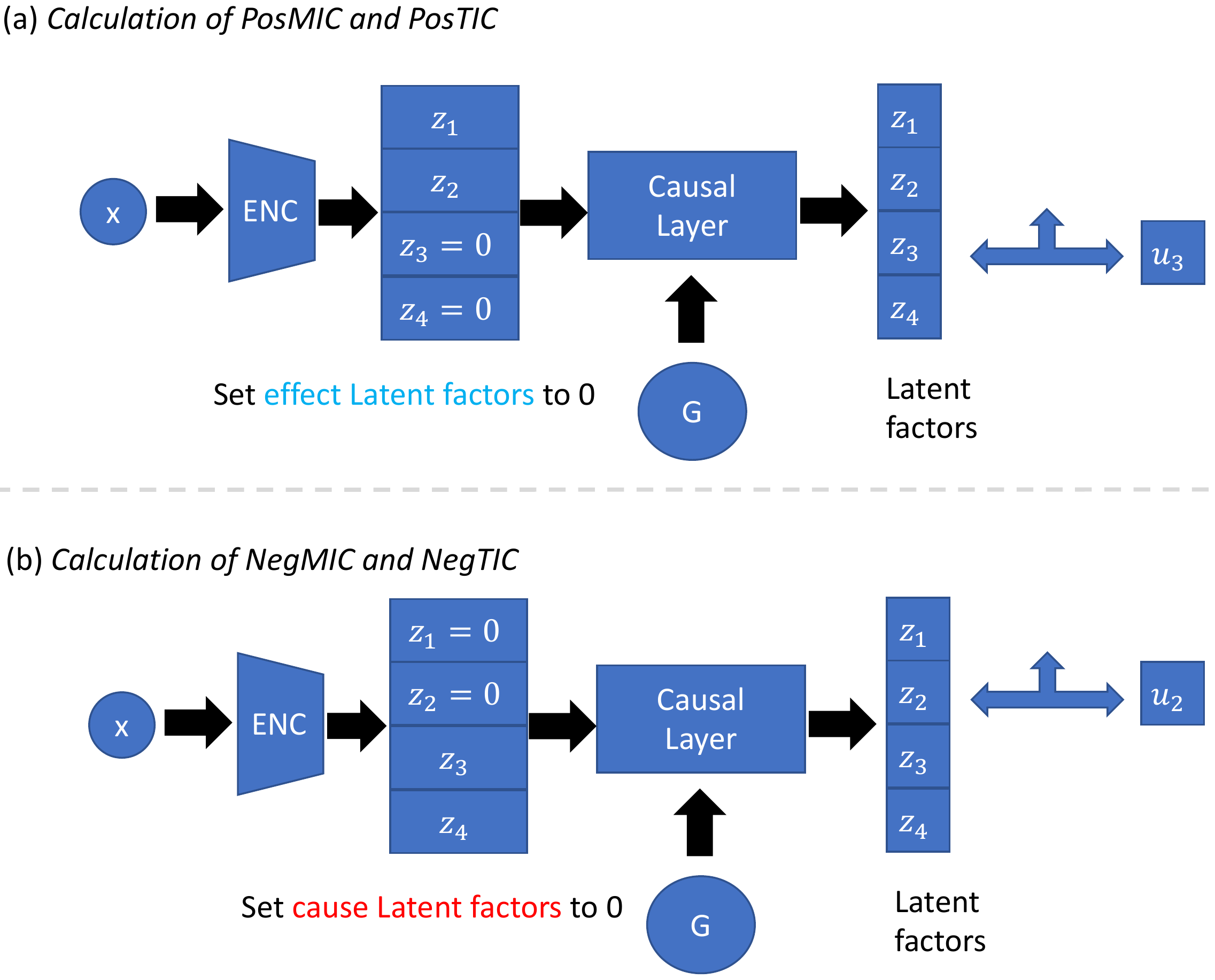}
% \vspace{-.5cm}
   \caption{One simple example of calculating new metrics.}
\label{fig:metric-calculation}
% \vspace{-.6cm}
\vspace{-\baselineskip}
\end{wrapfigure}
CausalVAE \cite{DBLP:conf/cvpr/YangLCSHW21} suggested using MIC and TIC for evaluating causal representation learning, despite the following inadequacy. In CausalVAE, MIC and TIC first calculate the information relevance between every ground truth labels and their corresponding learned latent factors. Then, the means of MIC and TIC for every factors are used as the final metrics values. However, MIC and TIC only measure correlations between a latent factor and its corresponding generative factor, and can not evaluate the correctness of relationships between cause and effect factors. Therefore, we argue that MIC and TIC are not suitable for evaluating causal representation learning where the goal is to learn the correct causal relationships between cause and effect factors. An intuitive example for illustrating the deficiency of MIC and TIC can be found in Appendix.

To address this issue, we propose four new metrics: PosMIC, PosTIC, NegMIC and NegTIC. PosMIC and PosTIC are used to evaluate the causal relation correctness  between latent factors, where higher value are expected. NegMIC and NegTIC are used to evaluate the falseness of causal relation discovery among latent factors, where lower value are expected. Additionally, to fully characterize the performance of causal representation learning using a single metric, we propose using the harmonic mean of the new metrics, i.e. $F_1^{MIC}$ and $F_1^{TIC}$. We will first describe how the proposed new metrics are calculated and then discuss their adequacy over the metrics used in CausalVAE.

\textbf{Calculating PosMIC, PosTIC, NegMIC and NegTIC:} As illustrated in \Cref{fig:metric-calculation}, to calculate PosMIC and PosTIC, given ground truth causal graph $G$, we first set the latent effect factors ($z_3$ and $z_4$ in \Cref{fig:metric-calculation}) to $0$. If the causal layer learns the correct relationship between the latent cause factors and the latent effect factors, $z_3$ and $z_4$ values are determined by the cause factors $z_1$ and $z_2$. Then, we separately calculate the MIC/TIC values of the latent effect factors and their corresponding generative factors. Finally, the means of the MIC/TIC of all latent effect factors values are taken to be the  PosMIC and PosTIC values. NegMIC and NegTIC are calculated in the opposite way, where the latent cause factors are set to $0$, and the final MIC/TIC values are calculated between the latent cause factors after the causal layer and their corresponding generative factors.  Ideally, the causal relationship should unidirectionally propagates from cause to effect, not in the opposite direction.  Thus, the lower NegMIC and NegTIC indicate better performance of causal representation learning.
To better compare different models and fully characterize the performance of causal representation learning, we consider Pos and Neg metrics together by calculating the harmonic mean: $F_1^{MIC} = 2 * \frac{PosMIC \cdot (1-NegMIC)}{PosMIC + (1-NegMIC)}$. $F_1^{TIC}$ of PosTIC and NegTIC is calculated similarly.

\begin{wrapfigure}{r}{0.4\textwidth}
\vspace{-\baselineskip}
\centering
%\fbox{\rule{0pt}{2in}
%\rule{0.9\linewidth}{0pt}}
  \includegraphics[width=0.4\textwidth]{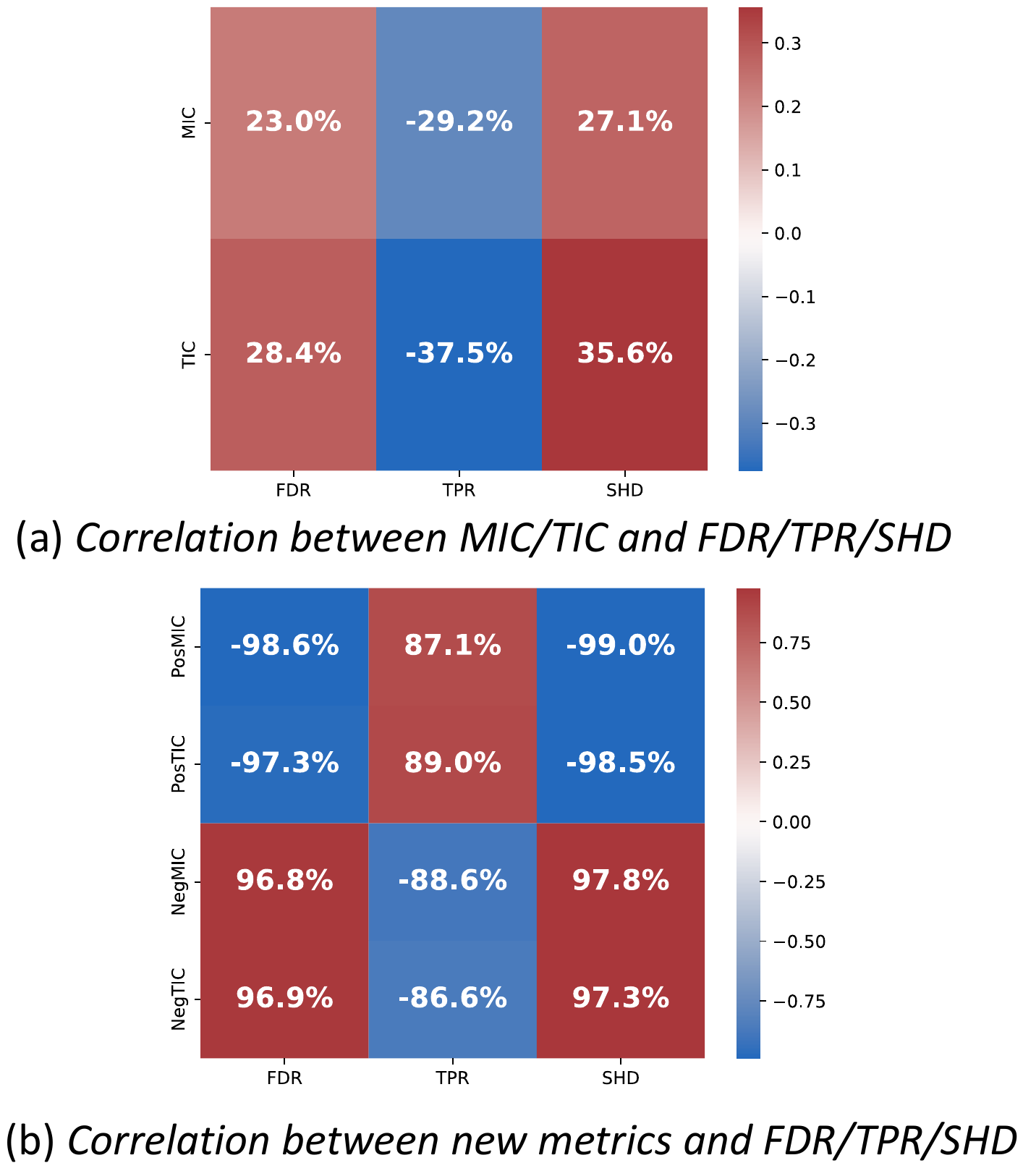}
% \vspace{-.5cm}
  \caption{Correlation of different metrics on Pendulum dataset. MIC and TIC show low correlation with rubrics for causal discovery. Contrarily, our proposed metrics shows high and expected correlation with those rubrics.}
\label{fig:metric-correlation}
% \vspace{-.6cm}
\vspace{-\baselineskip}
\end{wrapfigure}

\textbf{Adequacy of proposed metrics:} 
By conducting experiments on the Pendulum dataset, introduced in \Cref{sec:evaluation}, we empirically show the advantage of the new metrics  by proving that MIC and TIC fail to distinguish between models with correct and wrong causal graphs.  We initialize causal graphs $A$ of the CausalVAEs with different causal graphs and stop the gradient of elements if they are initialized with zero, such that CausalVAEs are created with various correctness levels of the causal graphs. If a causal graph $A$ is initialized identical to the correct causal graph, the performance of that CausalVAE is expected to be optimal since the correct causal relationship is obtained by initialization. Conversely, the performance of wrong causal graph initialized CausalVAE is expected to be poor. After training, we calculate correlations among metrics used for causal representation learning and rubrics used in the causal discovery research area: True Positive Rate (TPR), False Discovery Rate (FDR), and Structural Hamming Distance (SHD).  
TPR and FDR calculate the rate of discovering correct and wrong causal relations, respectively. SHD is the minimum number of modifications to correct a causal graph. As shown in \Cref{fig:metric-correlation}, MIC and TIC have a low correlation with TPR, FDR, and SHD. In contrast, our proposed new metrics PosMIC, PosTIC, NegMIC, and NegTIC have significant higher correlation with three rubrics used in causal inference. PosMIC and PosTIC are more positively correlated with TPR, and NegMIC and NegTIC are positively correlated with FDR and SHD. 
 
\section{Experimental Evaluation}
\label{sec:evaluation}

\begin{table}[]
\caption{Causal representation metrics tested on Pendulum and Flow.  }
\label{table:synthetic}
\renewcommand{\arraystretch}{1.3}
{
\begin{adjustbox}{width=\textwidth}
\setlength{\tabcolsep}{0.3em}
\begin{tabular}{ccccccccccccccccc}
\hlineB{3}
\multicolumn{1}{c|}{\multirow{2}{*}{Models}} & \multicolumn{8}{c|}{Pendulum}                                                                                                                                                                                           & \multicolumn{8}{c}{Flow}                                                                                                                                                                             \\ \cline{2-17} 
\multicolumn{1}{c|}{}                        & MIC $\mathbf{\uparrow}$                   & TIC  $\mathbf{\uparrow}$                 & PosMIC  $\mathbf{\uparrow}$               & PosTIC  $\mathbf{\uparrow}$               & NegMIC  $\mathbf{\downarrow}$               & NegTIC   $\mathbf{\downarrow}$               & $F_1^{MIC}$ $\mathbf{\uparrow}$                & \multicolumn{1}{c|}{$F_1^{TIC}$ $\mathbf{\uparrow}$}                  & MIC    $\mathbf{\uparrow}$                & TIC   $\mathbf{\uparrow}$                 & PosMIC $\mathbf{\uparrow}$                & PosTIC $\mathbf{\uparrow}$                & NegMIC $\mathbf{\downarrow}$                  & NegTIC  $\mathbf{\downarrow}$                & $F_1^{MIC}$ $\mathbf{\uparrow}$                  & $F_1^{TIC}$  $\mathbf{\uparrow}$                \\ \hline
\multicolumn{17}{c}{Fully Supervised learning methods (all labels are used)}                                                                                                                                                                                                                                                                                                                                                                                                  \\ \hline
\multicolumn{1}{c|}{CausalVAE}               & \textbf{95.1$\pm$2.1} & \textbf{81.6$\pm$1.9} & 53.0$\pm$4.5           & 43.4$\pm$3.7           & 46.6$\pm$3.9           & 37.0$\pm$4.2          & 53.2$\pm$3.6           & \multicolumn{1}{c|}{51.4$\pm$3.2}           & 72.1 $\pm$1.3          & 56.4 $\pm$1.6          & 45.1 $\pm$4.8          & 36.7 $\pm$4.2          & 43.3 $\pm$5.1          & 33.7 $\pm$3.2         & 50.2 $\pm$4.4          & 47.3 $\pm$3.7          \\ 
\multicolumn{1}{c|}{ConditionVAE}            & 93.8$\pm$3.3          & 80.5$\pm$1.4          & 36.5$\pm$3.0           & 27.8$\pm$3.2           & 34.6$\pm$4.2           & 25.7 $\pm$3.6         & 46.9 $\pm$4.7          & \multicolumn{1}{c|}{40.5 $\pm$3.5}          & \textbf{75.5 $\pm$2.3} & \textbf{56.5 $\pm$1.8} & 28.6 $\pm$3.2          & 21.3 $\pm$3.1          & 27.2 $\pm$2.8          & 20.6 $\pm$2.7         & 41.1 $\pm$5.1          & 33.6 $\pm$4.0          \\ \hline
\multicolumn{17}{c}{Unsupervised  Learning methods (no label is used)}                                                                                                                                                                                                                                                                                                                                                                                                        \\ \hline
\multicolumn{1}{c|}{CausalVAE(unsup)}        & 21.2 $\pm$1.4         & 12.0 $\pm$1.0         & 20.5 $\pm$2.6          & 11.8 $\pm$2.7          & 23.3 $\pm$3.2          & 14.7 $\pm$1.9         & 32.4 $\pm$3.4          & \multicolumn{1}{c|}{20.7 $\pm$3.1}          & 20.5 $\pm$4.7          & 11.8 $\pm$2.6          & 22.8 $\pm$2.7          & 12.5 $\pm$1.4          & 21.5 $\pm$2.4          & 12.0 $\pm$1.9         & 35.3 $\pm$5.6          & 21.9 $\pm$4.7          \\ 
\multicolumn{1}{c|}{BetaVAE}                 & 22.6 $\pm$4.6         & 12.5 $\pm$2.2         & 21.2 $\pm$2.7          & 12.7 $\pm$2.9          & 23.7 $\pm$3.1          & 12.6 $\pm$1.9         & 33.2 $\pm$3.3          & \multicolumn{1}{c|}{22.2 $\pm$2.7}          & 23.6 $\pm$3.2          & 12.5 $\pm$0.6          & 23.6 $\pm$3.6          & 12.5 $\pm$1.9          & 22.1 $\pm$2.5          & 11.4 $\pm$1.9         & 36.2 $\pm$4.9          & 21.9 $\pm$4.2          \\ 
\multicolumn{1}{c|}{LadderVAE}               & 22.4 $\pm$3.1         & 12.8 $\pm$1.2         & 15.2 $\pm$1.9          & 8.6 $\pm$1.0           & \textbf{14.2 $\pm$1.7} & \textbf{7.9 $\pm$0.9} & 25.8 $\pm$3.0          & \multicolumn{1}{c|}{15.7 $\pm$2.8}          & 34.3 $\pm$4.3          & 24.4 $\pm$1.5          & 16.2 $\pm$1.8          & 10.5 $\pm$1.0          & \textbf{13.3 $\pm$1.2} & \textbf{6.9 $\pm$0.6} & 27.3 $\pm$ 3.2         & 18.9 $\pm$2.8          \\ \hline
\multicolumn{17}{c}{Reduced supervision method (no label is used; supervision source is image pairing )}                                                                                                                                                                                                                                                                                                                                                                      \\ \hline
\multicolumn{1}{c|}{Our method}              & 86.6 $\pm$7.9         & 74.5 $\pm$5.1         & \textbf{54.1 $\pm$4.5} & \textbf{44.0 $\pm$4.2} & 40.2 $\pm$3.9          & 31.6 $\pm$3.2         & \textbf{56.8 $\pm$5.2} & \multicolumn{1}{c|}{\textbf{53.6 $\pm$4.3}} & 65.5 $\pm$6.6          & 56.7 $\pm$4.9          & \textbf{50.7 $\pm$4.7} & \textbf{41.3 $\pm$4.2} & 36.8 $\pm$3.8          & 27.2 $\pm$3.0         & \textbf{56.3 $\pm$5.9} & \textbf{52.7 $\pm$4.9} \\ \hlineB{3}
\end{tabular}

\end{adjustbox}
}
\vspace{-\baselineskip}
\end{table}

\textbf{Datasets:} Following \cite{DBLP:conf/cvpr/YangLCSHW21}, we use two synthetic datasets and two real world datasets.  \textbf{Pendulum} focuses on pendulum angle, light angle, shadow location and shadow length, and \textbf{Flow} focuses on ball size, water height, hole and water flow.
\textbf{CelebA(SMILE)} focuses on gender, smile, eyes open and mouth open, and \textbf{CelebA(BEARD)} focuses on age, gender and beardedness and baldness. We refer readers to \cite{DBLP:conf/cvpr/YangLCSHW21} and Appendix for more details. Besides using MIC and TIC as evaluation metrics, we also use our new metrics for better evaluating causal representation learning.

\subsection{Comparisons with State-Of-The-Art (SOTA)}

\textbf{Synthetic datasets:} 
Our method achieves comparable results on MIC and TIC compared with the fully supervised learning methods CausalVAE \cite{DBLP:conf/cvpr/YangLCSHW21} and ConditionVAE \cite{bib:conditionvae}, and outperform other unsupervised learning methods. As shown in \Cref{table:synthetic}, comparing to CausalVAE and ConditionVAE, our method can achieve slightly better performance on PosMIC, PosTIC, NegMIC and NegTIC. Unsupervised methods achieve low value on NegMIC and NegTIC due to barely learning semantic information.  Further, the result of using a few labels to train our method is included in Appendix.
\begin{figure}[]
    \centering % <-- added
\begin{subfigure}{0.21\textwidth}
  \includegraphics[width=0.90\textwidth]{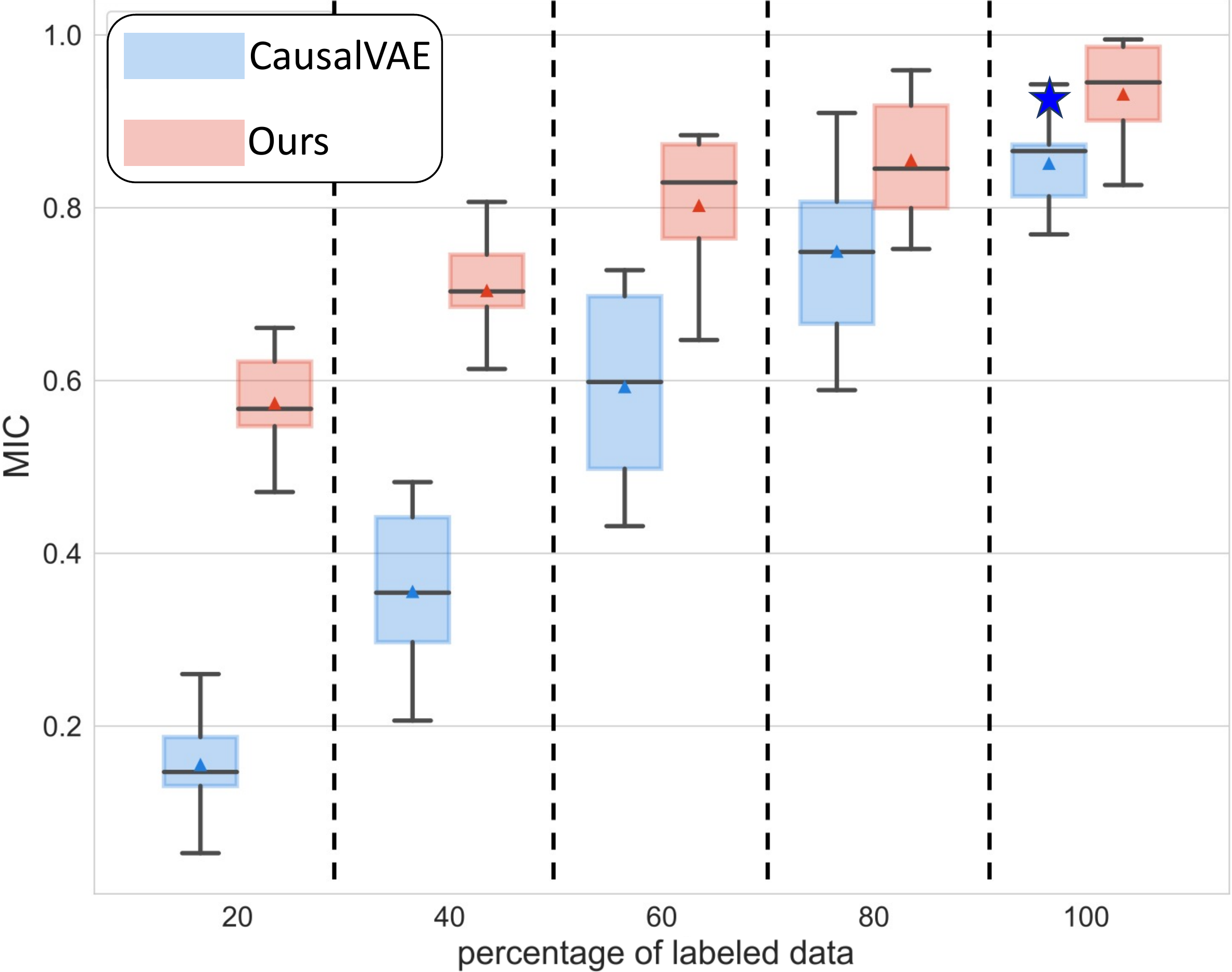}
  \caption{ MIC $\mathbf{\uparrow}$.}
  \label{fig:1}
\end{subfigure}\hfil % <-- added
\begin{subfigure}{0.21\textwidth}
  \includegraphics[width=0.90\textwidth]{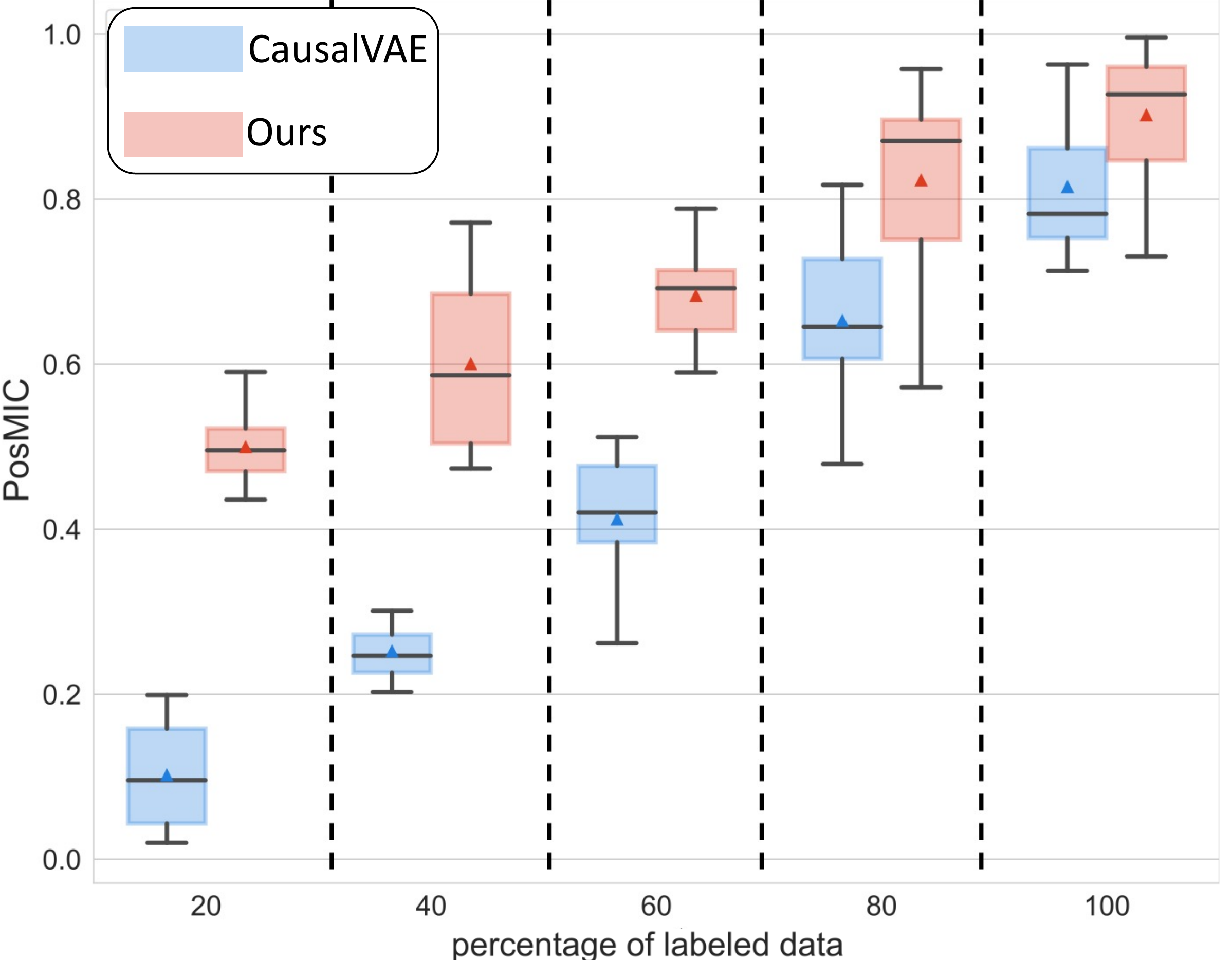}
  \caption{PosMIC $\mathbf{\uparrow}$.}
  \label{fig:2}
\end{subfigure}\hfil % <-- added
\begin{subfigure}{0.21\textwidth}
  \includegraphics[width=0.90\textwidth]{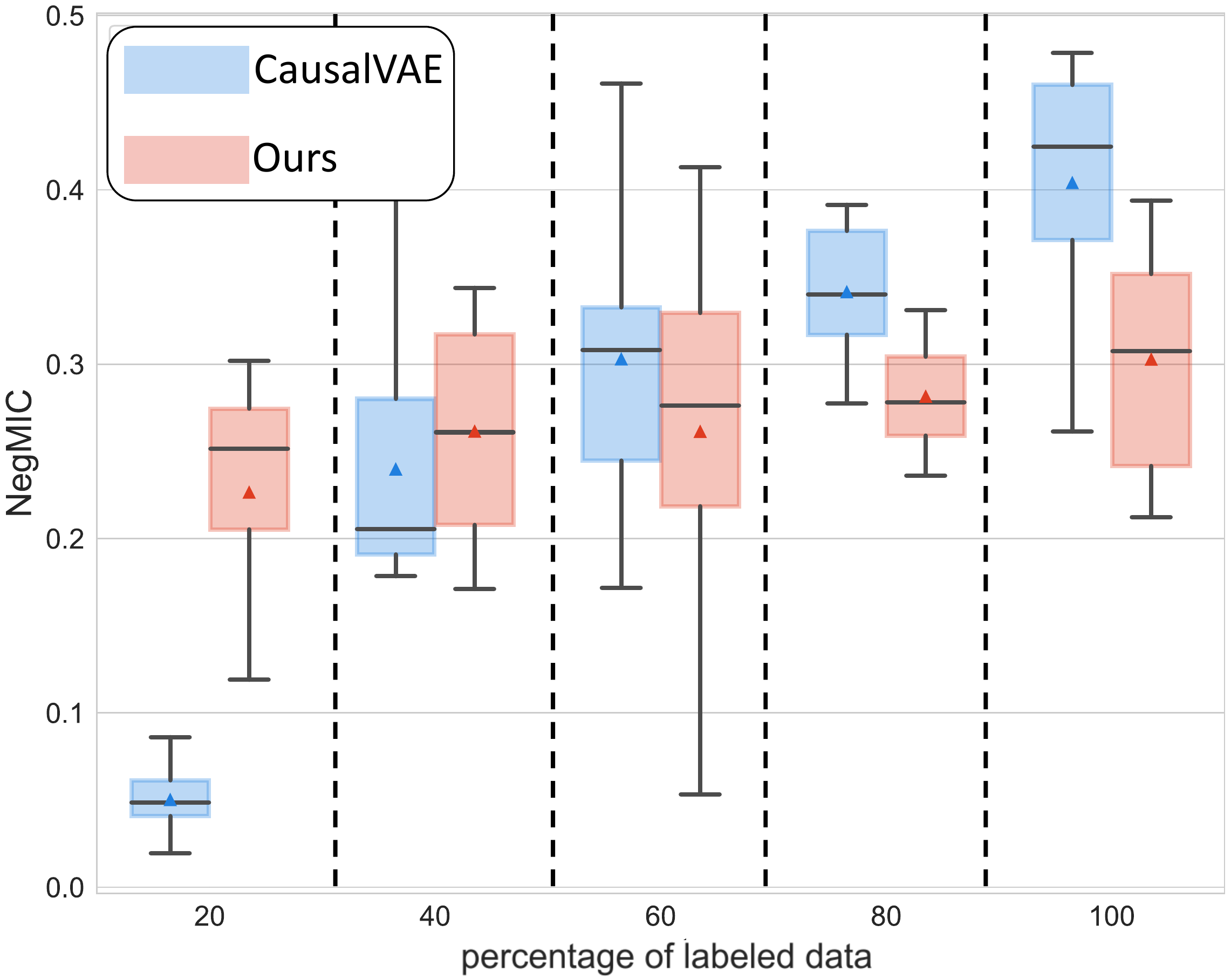}
  \caption{NegMIC $\mathbf{\downarrow}$.}
  \label{fig:3}
\end{subfigure} \hfil
\begin{subfigure}{0.21\textwidth}
  \includegraphics[width=0.90\textwidth]{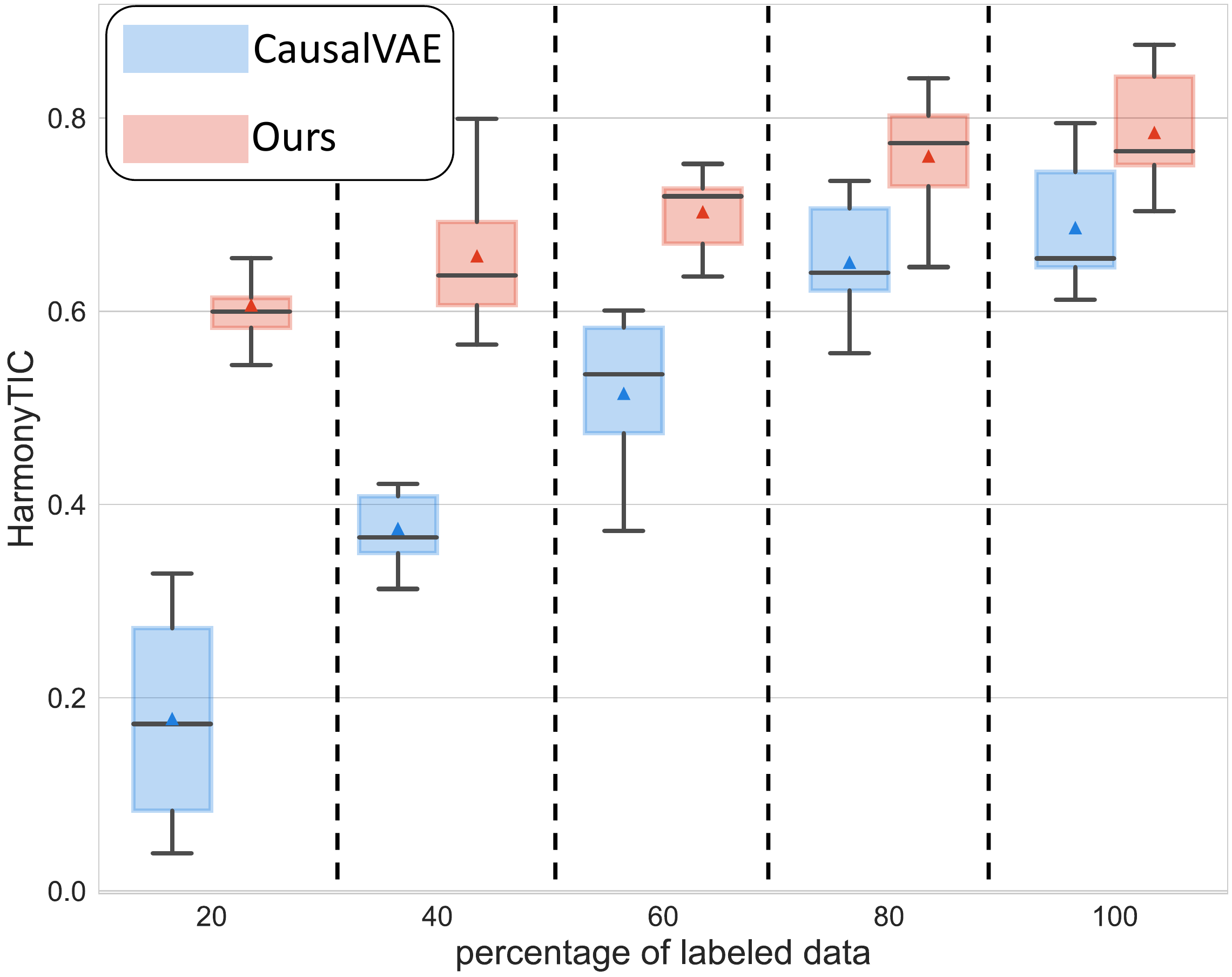}
  \caption{$F_1^{MIC}$ $\mathbf{\uparrow}$.}
  \label{fig:celeb_beard_f1mic}
\end{subfigure}

\medskip
\begin{subfigure}{0.21\textwidth}
  \includegraphics[width=0.90\textwidth]{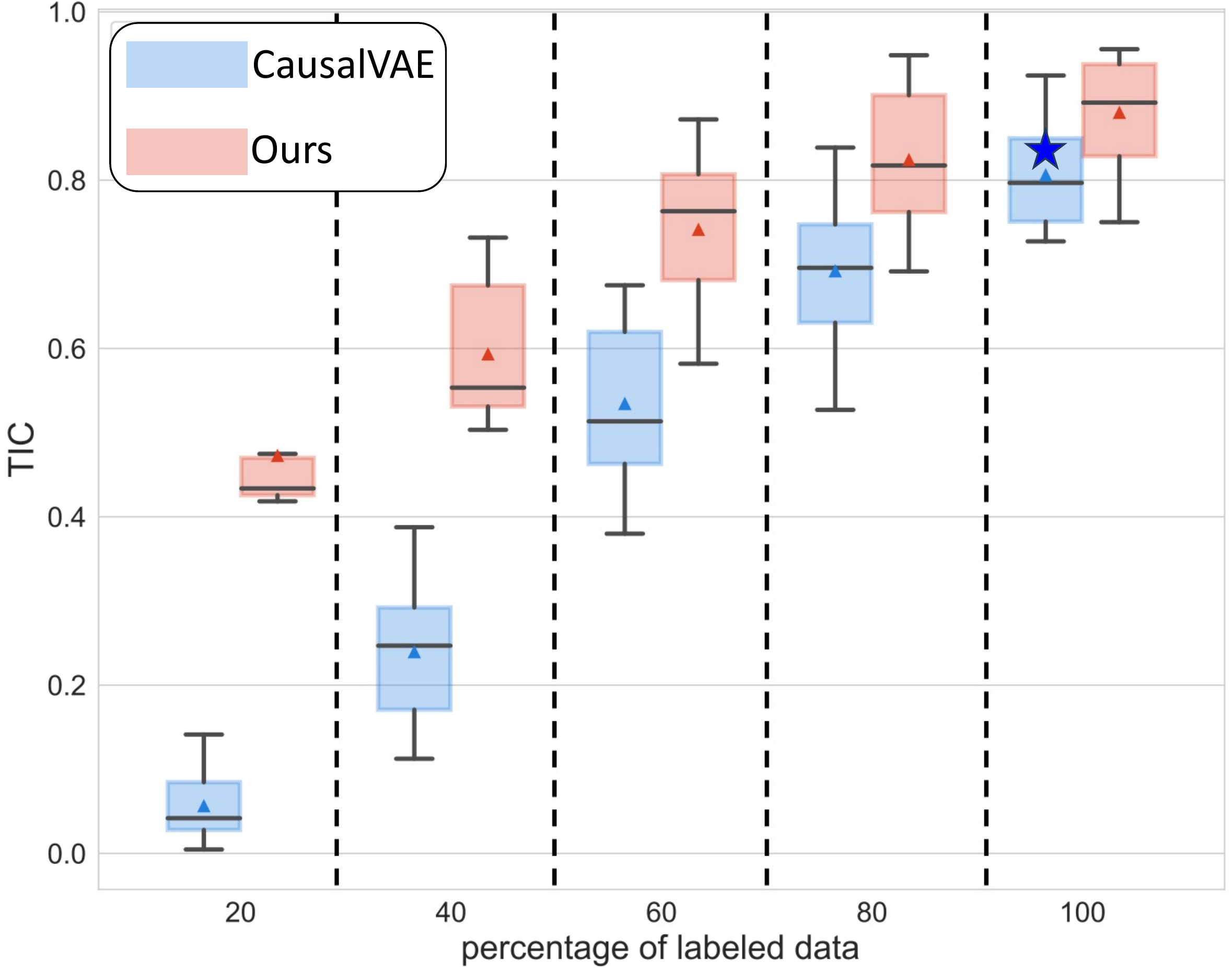}
  \caption{ TIC $\mathbf{\uparrow}$.}
  \label{fig:5}
\end{subfigure}\hfil % <-- added
\begin{subfigure}{0.21\textwidth}
  \includegraphics[width=0.90\textwidth]{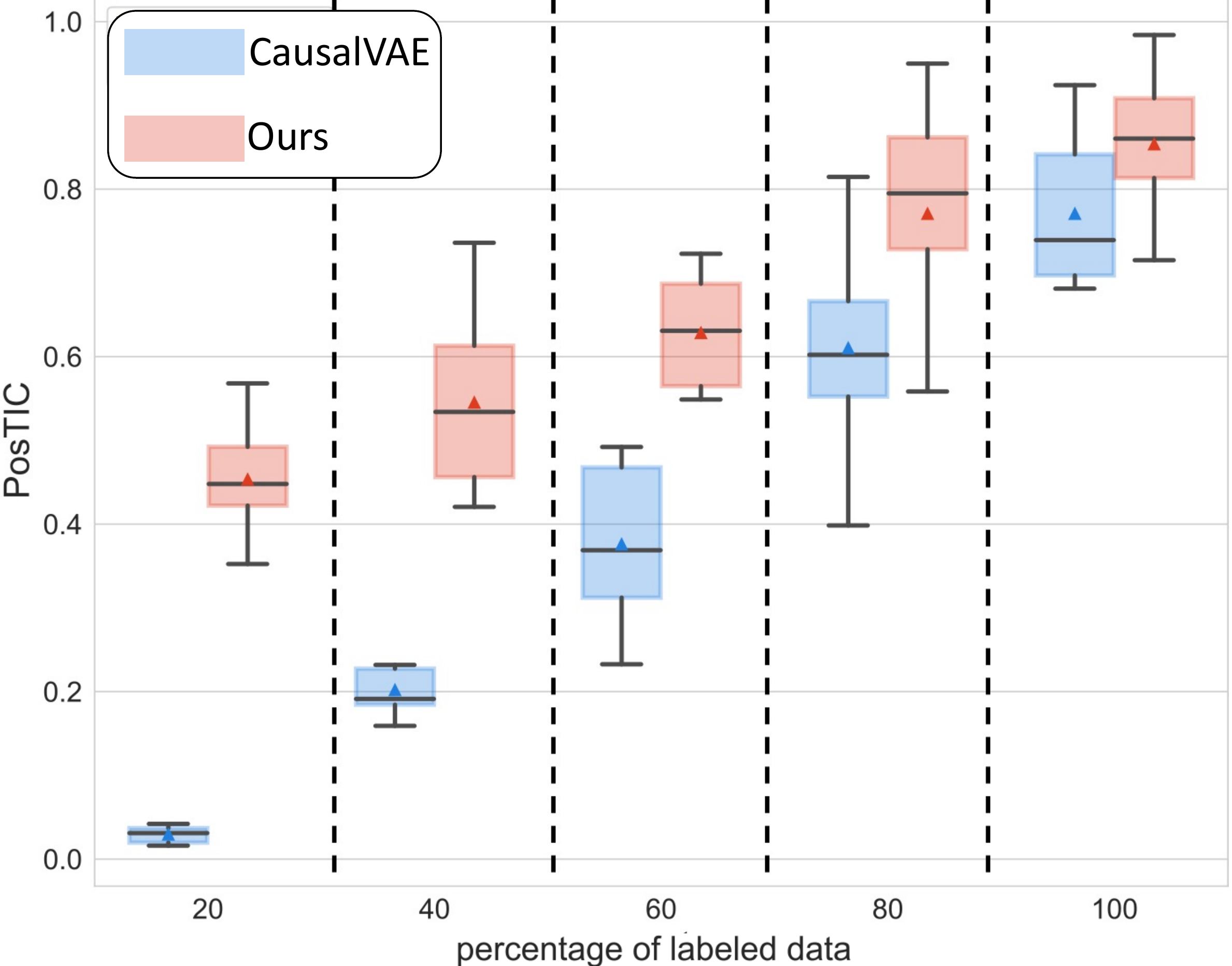}
  \caption{PosTIC $\mathbf{\uparrow}$.}
  \label{fig:6}
\end{subfigure}\hfil % <-- added
\begin{subfigure}{0.21\textwidth}
  \includegraphics[width=0.90\textwidth]{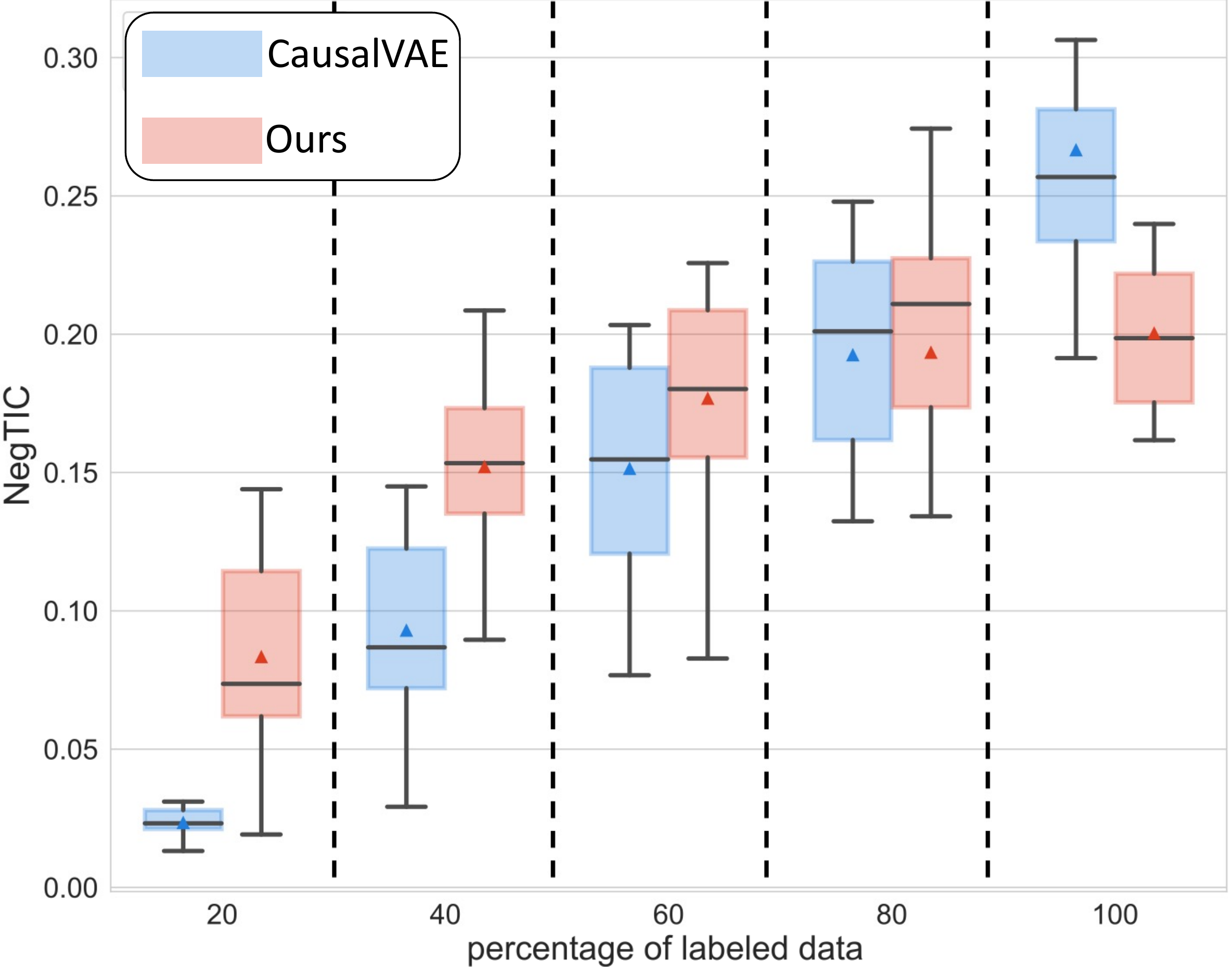}
  \caption{NegTIC $\mathbf{\downarrow}$.}
  \label{fig:7}
\end{subfigure}\hfil
\begin{subfigure}{0.21\textwidth}
  \includegraphics[width=0.90\textwidth]{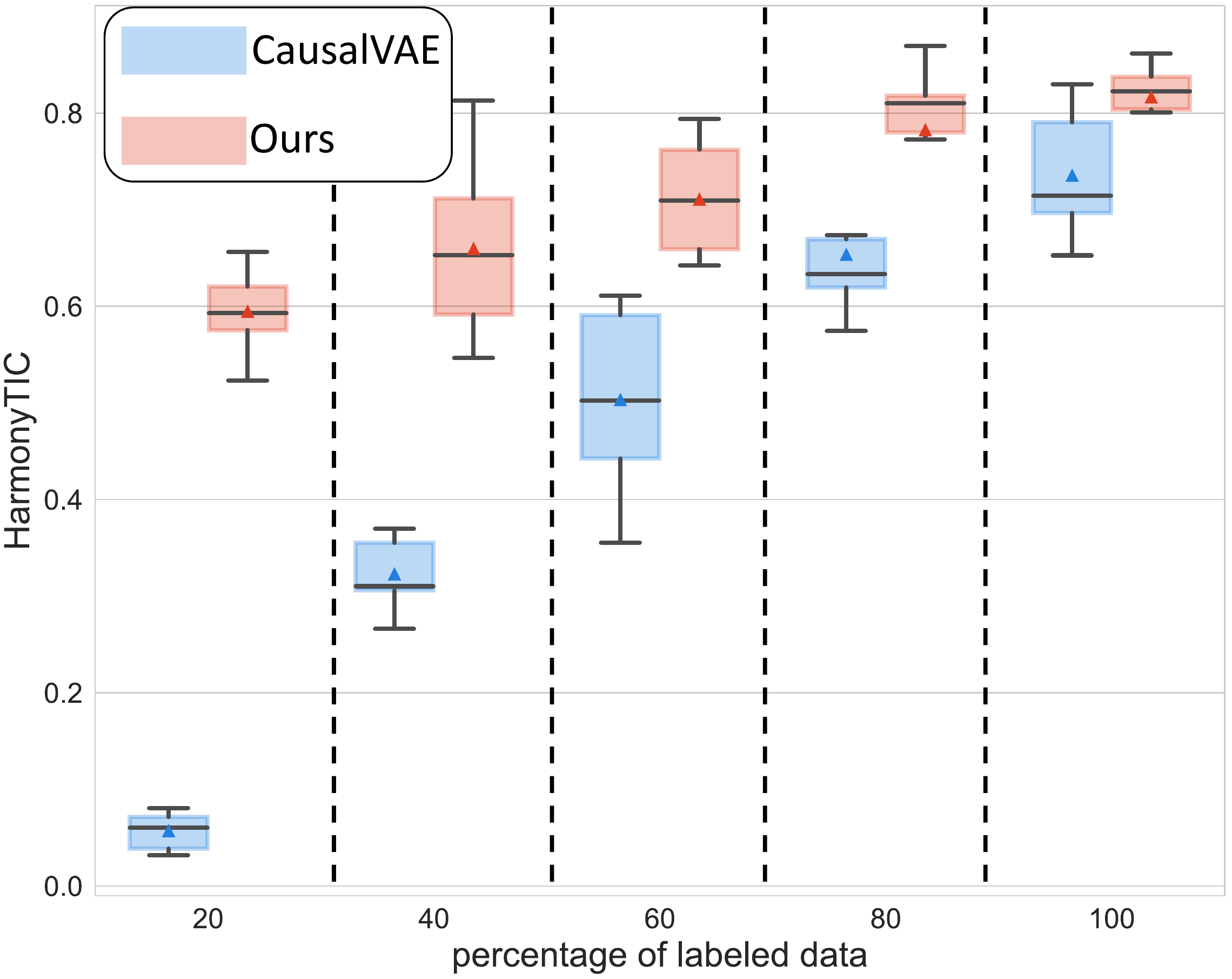}
  \caption{$F_1^{TIC}$ $\mathbf{\uparrow}$.}
  \label{fig:celeb_beard_f1tic}
\end{subfigure}
\caption{Box plots of metrics tested on CelebA(BEARD). Our method outperforms CausalVAE under various supervision strengths, where the advantage of our method is better revealed with weaker supervision strength. All experiments results are reproduced by us, except the blue star is the mean value reported in \cite{DBLP:conf/cvpr/YangLCSHW21}.}
\label{fig:mic-beard}
\end{figure}

\begin{figure}[]
    \centering % <-- added
\begin{subfigure}{0.21\textwidth}
  \includegraphics[width=0.90\textwidth]{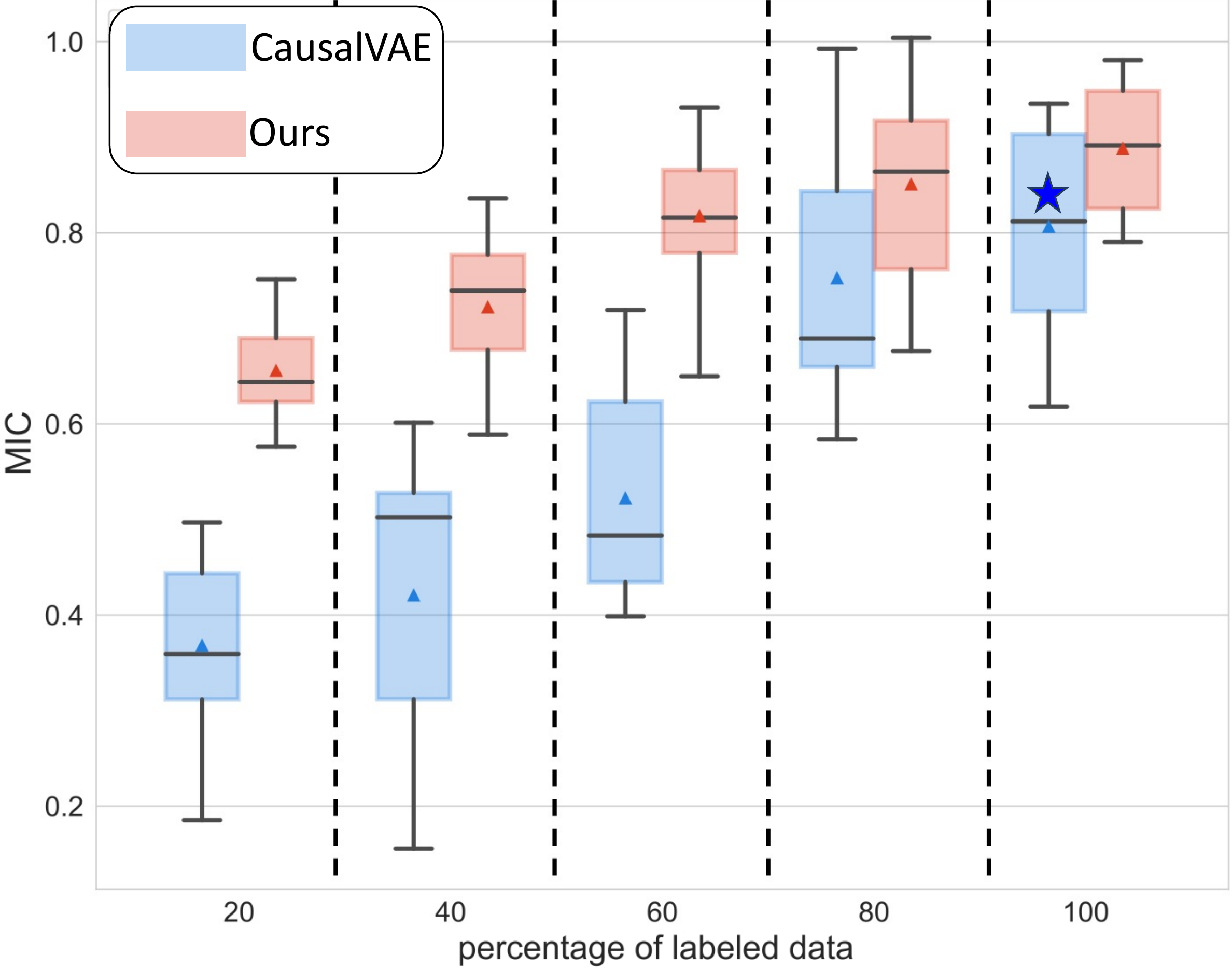}
  \caption{ MIC $\mathbf{\uparrow}$.}
  \label{fig:1}
\end{subfigure}\hfil % <-- added
\begin{subfigure}{0.21\textwidth}
  \includegraphics[width=0.90\textwidth]{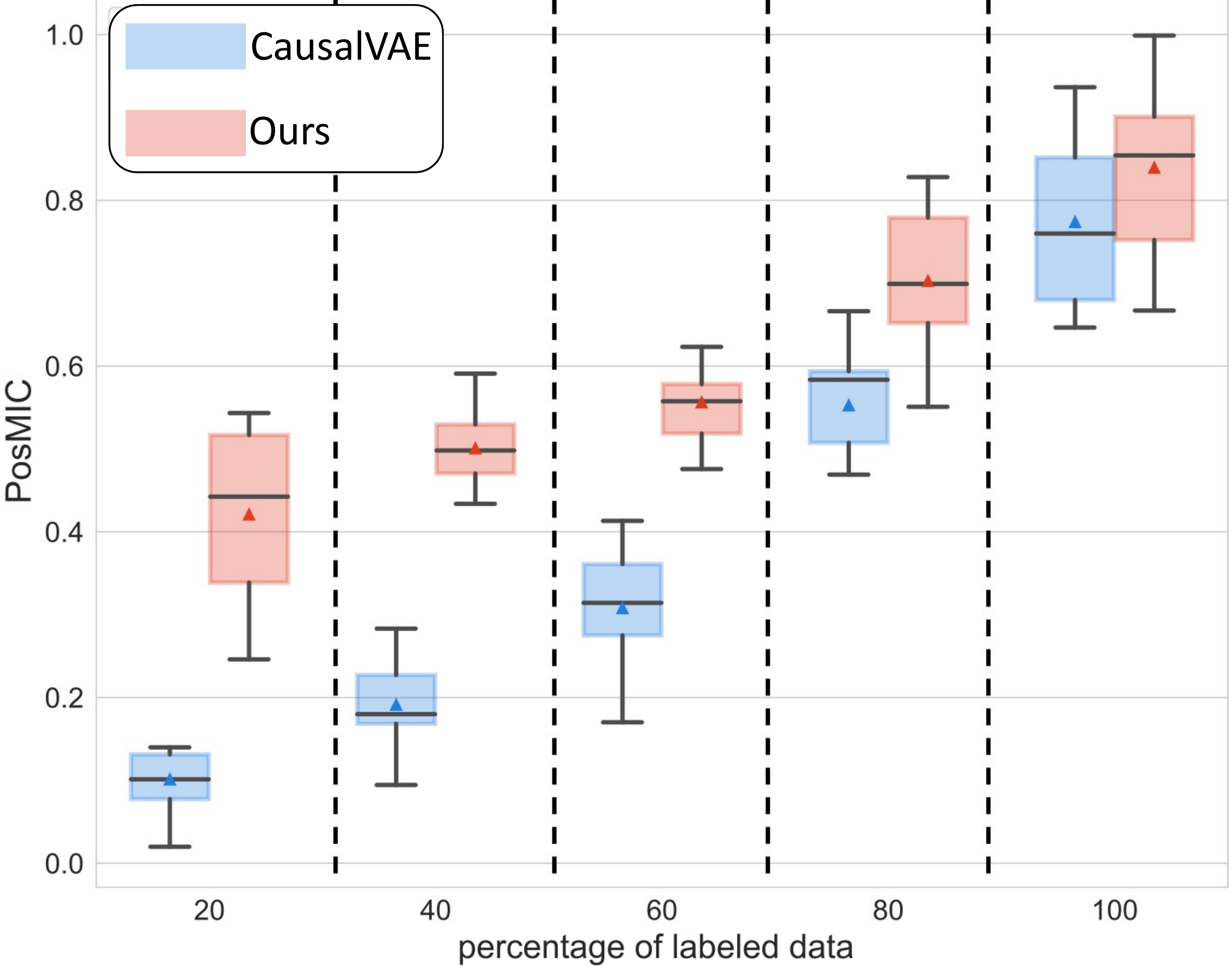}
  \caption{ PosMIC $\mathbf{\uparrow}$.}
  \label{fig:2}
\end{subfigure}\hfil % <-- added
\begin{subfigure}{0.21\textwidth}
  \includegraphics[width=0.90\textwidth]{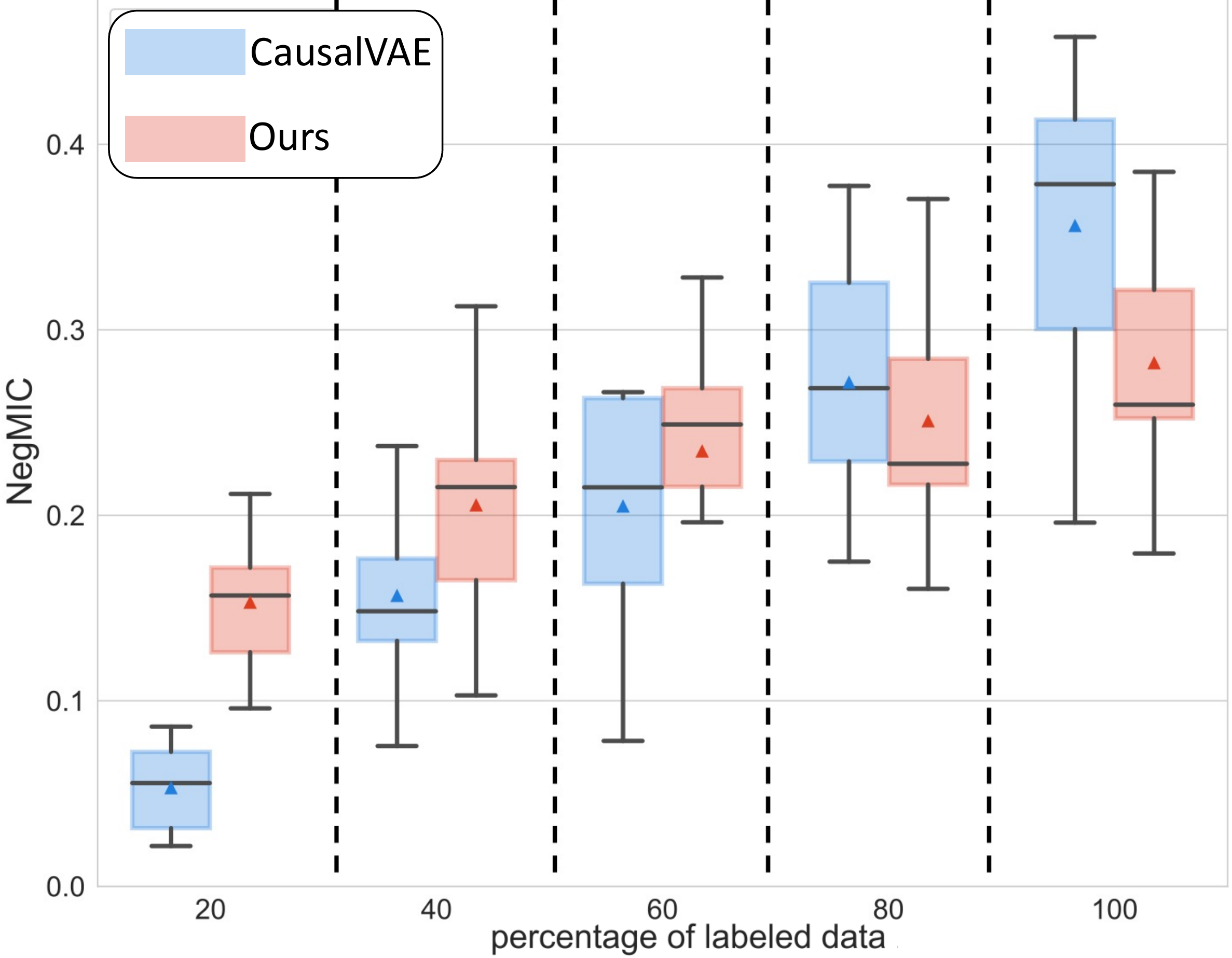}
  \caption{ NegMIC $\mathbf{\downarrow}$.}
  \label{fig:7}
\end{subfigure}\hfil
\begin{subfigure}{0.21\textwidth}
  \includegraphics[width=0.90\textwidth]{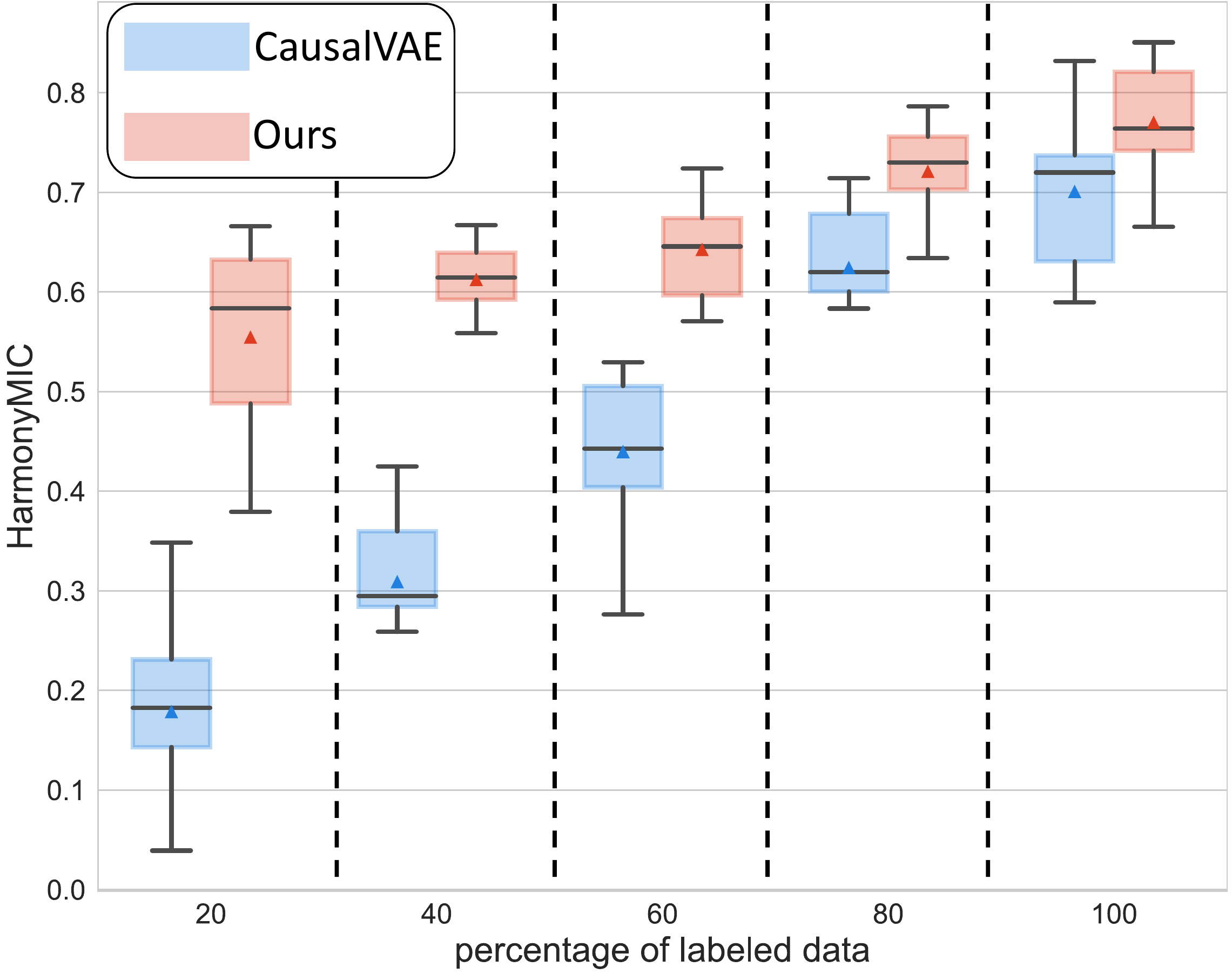}
  \caption{$F_1^{MIC}$ $\mathbf{\uparrow}$.}
  \label{fig:celeb_smile_f1mic}
\end{subfigure}

\medskip
\begin{subfigure}{0.21\textwidth}
  \includegraphics[width=0.90\textwidth]{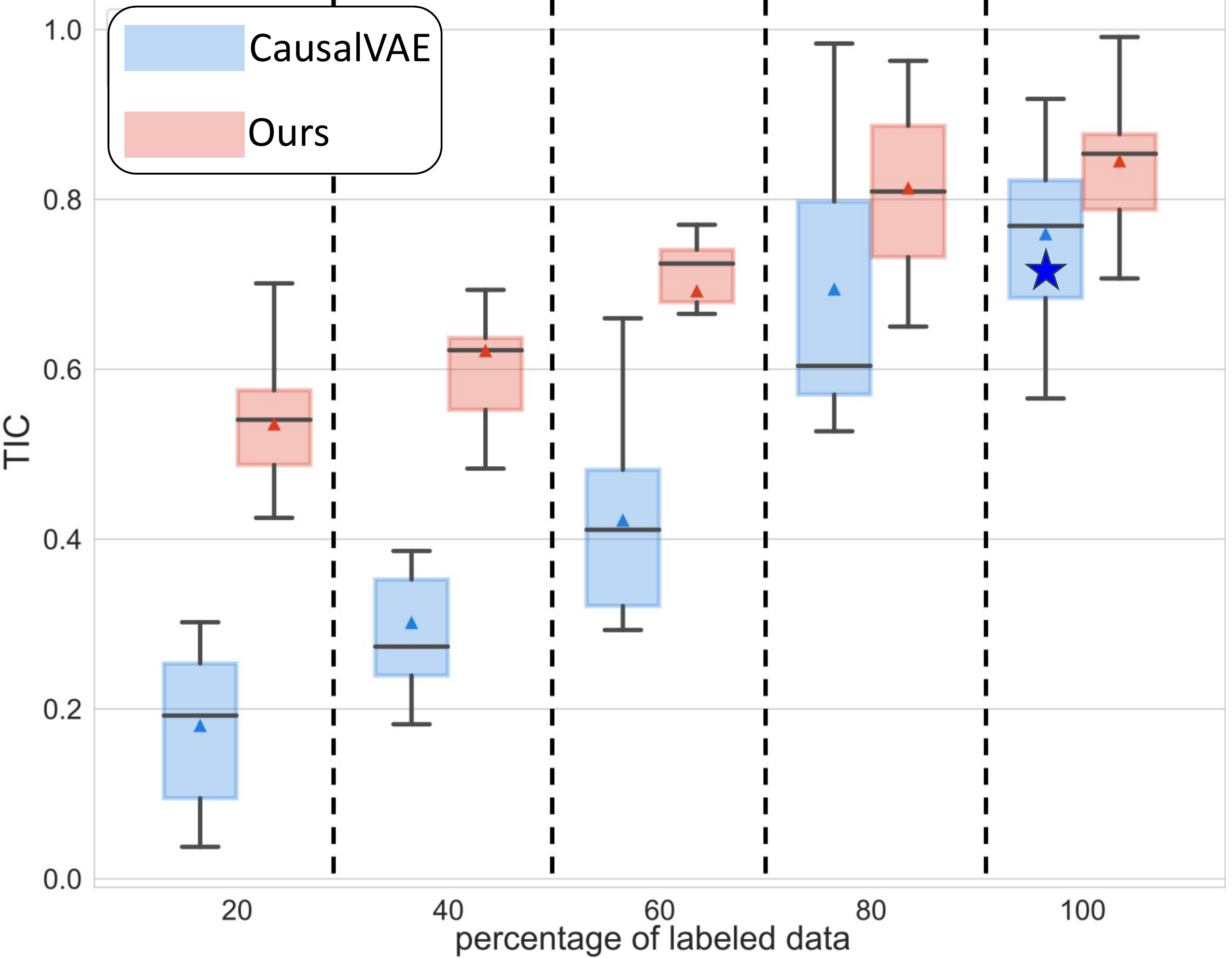}
  \caption{ TIC $\mathbf{\uparrow}$.}
  \label{fig:4}
\end{subfigure}\hfil % <-- added
\begin{subfigure}{0.21\textwidth}
  \includegraphics[width=0.90\textwidth]{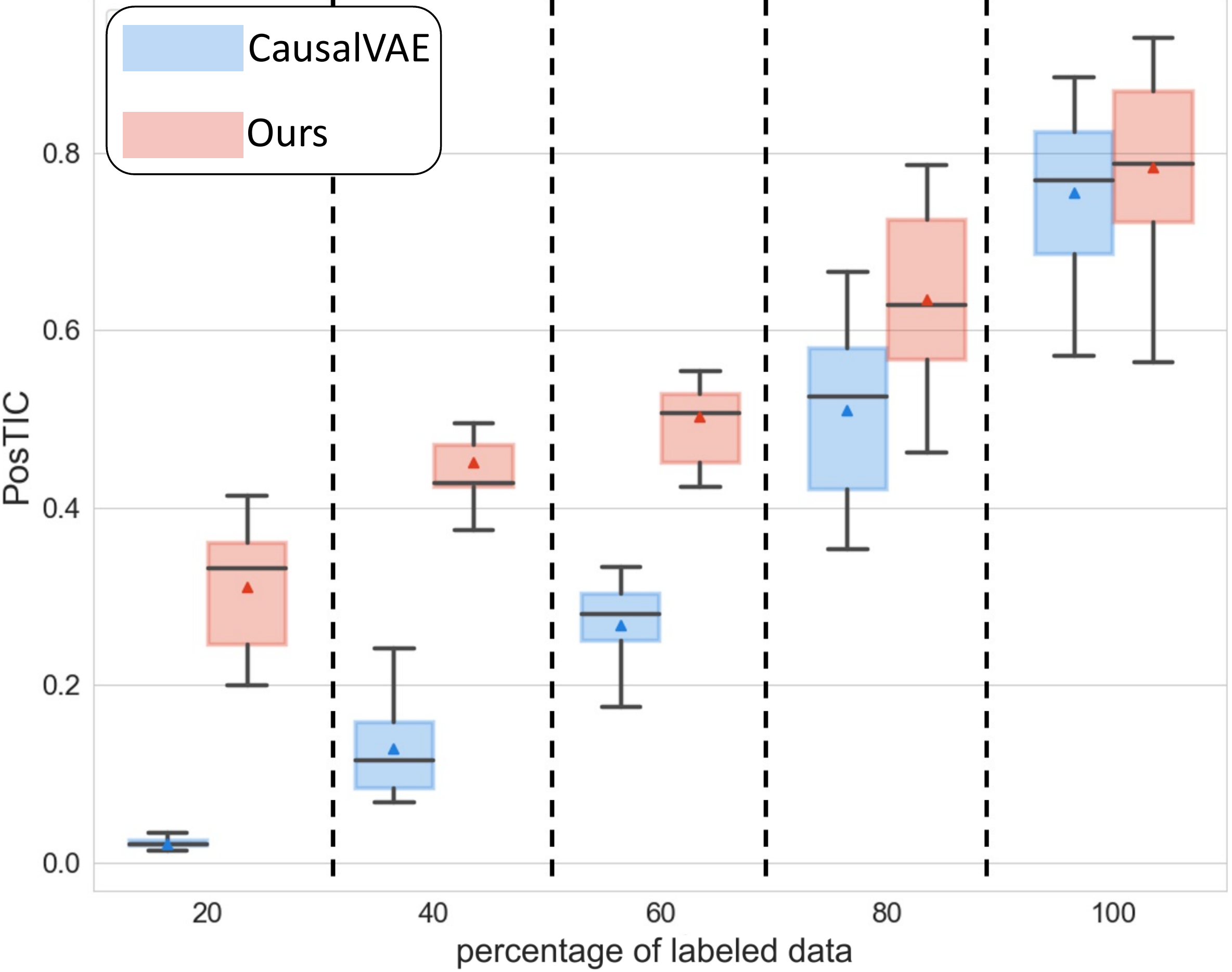}
  \caption{ PosTIC $\mathbf{\uparrow}$.}
  \label{fig:5}
\end{subfigure}\hfil % <-- added
\begin{subfigure}{0.21\textwidth}
  \includegraphics[width=0.90\textwidth]{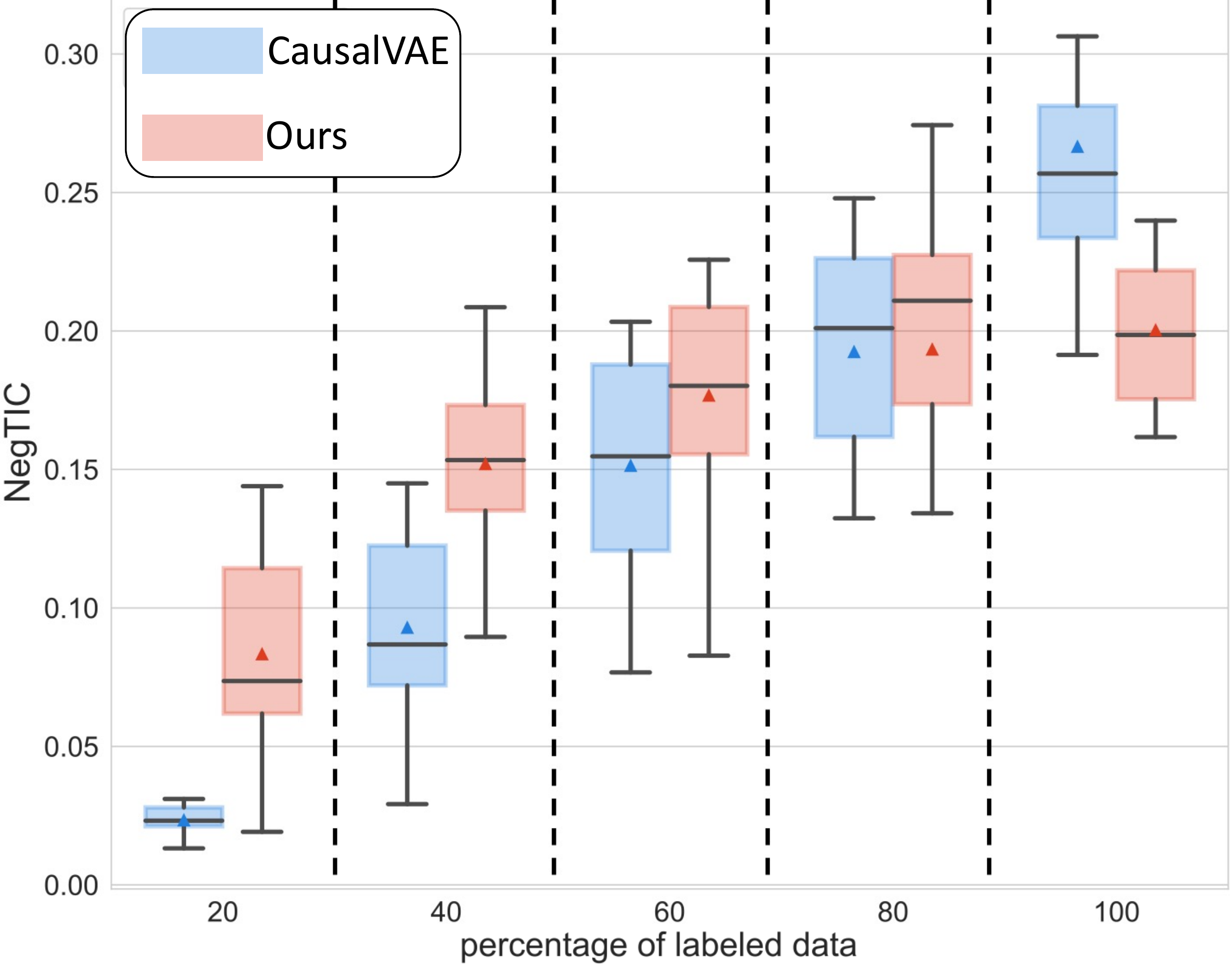}
  \caption{NegTIC $\mathbf{\downarrow}$.}
  \label{fig:6}
\end{subfigure}
\begin{subfigure}{0.211\textwidth}
  \includegraphics[width=0.90\textwidth]{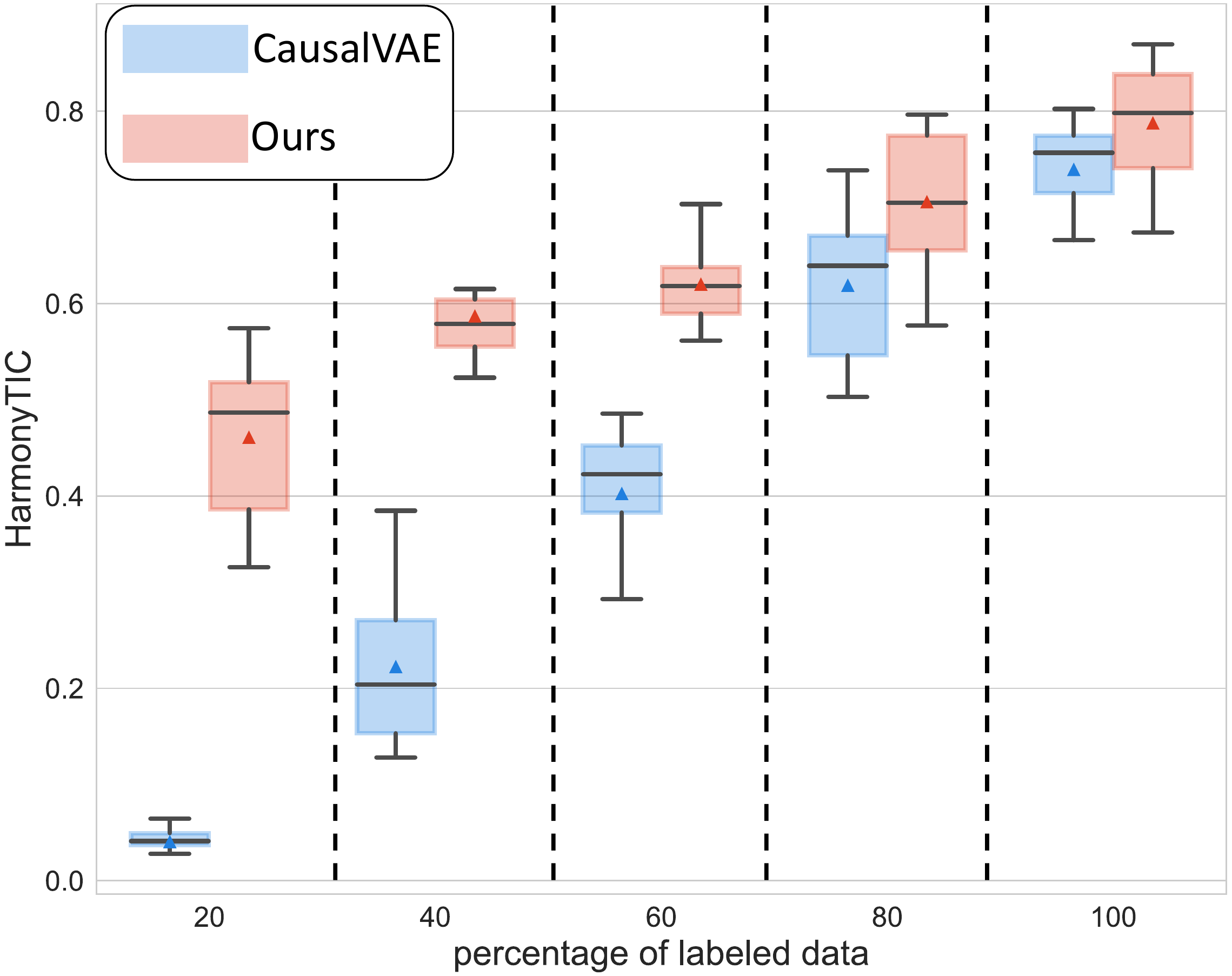}
  \caption{$F_1^{TIC}$ $\mathbf{\uparrow}$.}
  \label{fig:celeb_smile_f1tic}
\end{subfigure}
\caption{Box plots of metrics tested on CelebA(SMILE). Our method outperforms CausalVAE under various supervision strengths, where the advantage of our method is better revealed with weaker supervision strength. All experiments results are reproduced by us, except the blue star is the mean value reported in \cite{DBLP:conf/cvpr/YangLCSHW21}.}
\label{fig:mic-smile}
\end{figure}

\textbf{Real datasets:}
Compared with synthetic datasets, there are $40$ generative factors in CelebA dataset. If no label is available during training CelebA datasets, the search space for the model becomes intractable as there are $2^{40}$  different binary causal graphs for $40$ factors. To decrease the difficulty, label information is needed to control the semantic factors which are encoded in each dimension of latent space \cite{bib:dual-disentangle}. For comprehensive comparison, our model and baselines are trained with $\{20\%, 40\%, 60\%, 80\%, 100\% \}$ of labeled data, and the remaining samples are unlabelled. As shown in  \Cref{fig:mic-beard,fig:mic-smile}, our method consistently and significantly outperforms CausalVAE. Furthermore, with fewer labels, our method outperforms CausalVAE more appreciably. 

% Noticeably, when $20\%$ of labels are used, although CausalVAE shows low NegMIC and NegTIC, it fails to encode useful semantic information from inputs as all other metrics are dissatisfied. With stronger supervision, NegMIC and NegTIC of CausalVAE increase rapidly while our methods are relatively stable, indicating the stability of our method.

% \subsection{Balance between positive and negative metrics}
% As described in \Cref{sec:metric}, Pos and Neg metrics are both important for evaluating causal representation learning since PosMIC/TIC and NegMIC/TIC measure correct and false causal relationships discovery respectively. However, satisfying NegMIC/TIC can be easily reached if a model barely learns semantic information (\eg{} a randomly initialized model without training might achieve perfect results). To consider both Pos and Neg metrics, as discussed in \Cref{sec:metric}, we use $F_1^{MIC}$ and $F_1^{TIC}$ to integrate both assessment aspects. The results on synthetic datasets are shown in \Crefrange{table:pendulum}{table:flow}, and the results on real datasets are shown in \Cref{fig:celeb_beard_f1mic,fig:celeb_beard_f1tic} and \Cref{fig:celeb_smile_f1mic,fig:celeb_smile_f1tic}. Our method outperforms all SOTAs, which shows the advantage of our method when using reduced supervision strength.
% on learning correct causal relationships while preventing extracting false relationships, using reduced supervision strength.

% \subsection{Importance of the \emph{do-operation}}
% To prove the importance of the different \emph{do-operation} modules used in our method, we evaluate our model by removing different \emph{do-operation} modules in the architecture. As shown in  \Cref{table:ablation1}, by removing Do-Cause, the model loses the ability of finding causal relationship. Removing Do-effect will lead to performance decrease on NegMIC and NegTIC.

% Therefore, the performance on MIC, TIC, PosMIC and PosTIC degrades similar to unsupervised CausalVAE. 
% By removing Do-effect and keeping Do-Cause, 
% the performance on MIC, TIC, PosMIC and PosTIC significantly improves, while the performance on NegMIC and NegTIC is worse than full model where both cause and effect \emph{do-operation} modules are used.

% \input{tables/ablation}

\section{Conclusion}
\label{sec:conclusion}

In this work, we propose a novel architecture for causal representation learning with reduced supervision strength, exploiting the \emph{do-operation} . We use a pair of images and apply \emph{do-operation} to both latent cause and effect factors for new reconstructions. By comparing the new reconstructions after \emph{do-operation} and the original inputs, the supervision strength is reduced. Furthermore, to better evaluate causal representation learning, we propose new metrics to address adequacy of existing metrics. We empirically demonstrate the advantages of our method on both synthetic and real datasets.

\noindent\textbf{Acknowledgement:} This material is based on research sponsored by Air Force Research Laboratory under
agreement number FA8750-19-1-1000. The U.S. Government is authorized to reproduce and distribute reprints for Governmental purposes notwithstanding any copyright notation thereon.

\bibliography{iclr2023_conference}

\begin{thebibliography}{10}\itemsep=-1pt

\bibitem{bib:bengio_representaion}
Yoshua Bengio, Aaron Courville, and Pascal Vincent.
\newblock Representation learning: A review and new perspectives.
\newblock {\em IEEE Transactions on Pattern Analysis and Machine Intelligence},
  35(8):1798--1828, 2013.

\bibitem{burgess2018understanding}
Christopher~P. Burgess, Irina Higgins, Arka Pal, Loic Matthey, Nick Watters,
  Guillaume Desjardins, and Alexander Lerchner.
\newblock Understanding disentangling in $\beta$-vae, 2018.

\bibitem{chen2019isolating}
Ricky T.~Q. Chen, Xuechen Li, Roger Grosse, and David Duvenaud.
\newblock Isolating sources of disentanglement in variational autoencoders,
  2019.

\bibitem{bib:dual-disentangle}
Zunlei Feng, Xinchao Wang, Chenglong Ke, An-Xiang Zeng, Dacheng Tao, and Mingli
  Song.
\newblock Dual swap disentangling.
\newblock In S. Bengio, H. Wallach, H. Larochelle, K. Grauman, N. Cesa-Bianchi,
  and R. Garnett, editors, {\em Advances in Neural Information Processing
  Systems}, volume~31. Curran Associates, Inc., 2018.

\bibitem{DBLP:conf/iclr/HigginsMPBGBML17}
Irina Higgins, Lo{\"{\i}}c Matthey, Arka Pal, Christopher Burgess, Xavier
  Glorot, Matthew Botvinick, Shakir Mohamed, and Alexander Lerchner.
\newblock beta-vae: Learning basic visual concepts with a constrained
  variational framework.
\newblock In {\em 5th International Conference on Learning Representations,
  {ICLR} 2017, Toulon, France, April 24-26, 2017, Conference Track
  Proceedings}. OpenReview.net, 2017.

\bibitem{bib:ivae}
Ilyes Khemakhem, Diederik Kingma, Ricardo Monti, and Aapo Hyvarinen.
\newblock Variational autoencoders and nonlinear ica: A unifying framework.
\newblock In Silvia Chiappa and Roberto Calandra, editors, {\em Proceedings of
  the Twenty Third International Conference on Artificial Intelligence and
  Statistics}, volume 108 of {\em Proceedings of Machine Learning Research},
  pages 2207--2217. PMLR, 26--28 Aug 2020.

\bibitem{DBLP:journals/corr/KingmaW13}
Diederik~P. Kingma and Max Welling.
\newblock Auto-encoding variational bayes.
\newblock In Yoshua Bengio and Yann LeCun, editors, {\em 2nd International
  Conference on Learning Representations, {ICLR} 2014, Banff, AB, Canada, April
  14-16, 2014, Conference Track Proceedings}, 2014.

\bibitem{bib:mic-tic}
Justin~B. Kinney and Gurinder~S. Atwal.
\newblock Equitability, mutual information, and the maximal information
  coefficient.
\newblock {\em Proceedings of the National Academy of Sciences},
  111(9):3354--3359, 2014.

\bibitem{bib:laddervae}
Daniel~D. Lee, Masashi Sugiyama, Ulrike von Luxburg, Isabelle Guyon, and Roman
  Garnett, editors.
\newblock {\em Advances in Neural Information Processing Systems 29: Annual
  Conference on Neural Information Processing Systems 2016, December 5-10,
  2016, Barcelona, Spain}, 2016.

\bibitem{locatello2019challenging}
Francesco Locatello, Stefan Bauer, Mario Lucic, Gunnar Rätsch, Sylvain Gelly,
  Bernhard Schölkopf, and Olivier Bachem.
\newblock Challenging common assumptions in the unsupervised learning of
  disentangled representations, 2019.

\bibitem{Locatello2020Disentangling}
Francesco Locatello, Michael Tschannen, Stefan Bauer, Gunnar Rätsch, Bernhard
  Schölkopf, and Olivier Bachem.
\newblock Disentangling factors of variations using few labels.
\newblock In {\em International Conference on Learning Representations}, 2020.

\bibitem{bib:gae}
Ignavier Ng, Shengyu Zhu, Zhitang Chen, and Zhuangyan Fang.
\newblock A graph autoencoder approach to causal structure learning.
\newblock {\em arXiv preprint arXiv:1911.07420}, 2019.

\bibitem{reason:Pearl09a}
Judea Pearl.
\newblock {\em Causality: Models, Reasoning and Inference}.
\newblock Cambridge University Press, 2nd edition, 2009.

\bibitem{bib:towards-causal}
Bernhard Schölkopf, Francesco Locatello, Stefan Bauer, Nan~Rosemary Ke, Nal
  Kalchbrenner, Anirudh Goyal, and Yoshua Bengio.
\newblock Toward causal representation learning.
\newblock {\em Proceedings of the IEEE}, 109(5):612--634, 2021.

\bibitem{bib:conditionvae}
Kihyuk Sohn, Honglak Lee, and Xinchen Yan.
\newblock Learning structured output representation using deep conditional
  generative models.
\newblock In C. Cortes, N. Lawrence, D. Lee, M. Sugiyama, and R. Garnett,
  editors, {\em Advances in Neural Information Processing Systems}, volume~28.
  Curran Associates, Inc., 2015.

\bibitem{DBLP:conf/cvpr/YangLCSHW21}
Mengyue Yang, Furui Liu, Zhitang Chen, Xinwei Shen, Jianye Hao, and Jun Wang.
\newblock Causalvae: Disentangled representation learning via neural structural
  causal models.
\newblock In {\em {IEEE} Conference on Computer Vision and Pattern Recognition,
  {CVPR} 2021, virtual, June 19-25, 2021}, pages 9593--9602. Computer Vision
  Foundation / {IEEE}, 2021.

\bibitem{bib:notears}
Xun Zheng, Bryon Aragam, Pradeep Ravikumar, and Eric~P. Xing.
\newblock {DAGs with NO TEARS: Continuous Optimization for Structure Learning}.
\newblock In {\em Advances in Neural Information Processing Systems}, 2018.

\end{thebibliography}
\bibliographystyle{test}

\clearpage
\appendix

% to compile a preprint version, e.g., for submission to arXiv, add add the
% [preprint] option:
%     \usepackage[preprint]{neurips_2022}

% to compile a camera-ready version, add the [final] option, e.g.:
%     \usepackage[final]{neurips_2022}

% to avoid loading the natbib package, add option nonatbib:
%    \usepackage[nonatbib]{neurips_2022}

%%%%%%%%%%%%%%%%%%%%%%%%%%%%%%%%%%%%%%%%%%%%%%%%%%%%%%%%%%%%

\setcounter{table}{0}
\renewcommand{\thetable}{A\arabic{table}}
\setcounter{figure}{0}
\renewcommand{\thefigure}{A\arabic{figure}}

\section{Appendix}
% \subsection{Ablation Study}
% To prove the importance of different \emph{do-operation} module used in our method, we experiments our model by removing different \emph{do-operation} modules in the model. As shown in  \Cref{table:ablation1}, by removing cause \emph{do-operation} module, the model lose ability of finding causal relationship. Therefore, the performance on MIC, TIC, PosMIC and PosTIC are similar to unsupervised CausalVAE. By removing effect \emph{do-operation} and keeping cause \emph{do-operation}, the the performance on MIC, TIC, PosMIC and PosTIC get significant improvement but performance on NegMIC and NegTIC are still worse than baseline model which both cause and effect \emph{do-operation} modules are used.
% \begin{table}[!htp]
% \centering
% \caption{Causal representation metrics of model with different \emph{do-operation} module applied}
% \label{table:ablation1}
% \begin{adjustbox}{width=\textwidth}
% \begin{tabular}{ll|llllll}
% \hline
% Cause Do-operation            & Effect Do-operation            & MIC  & TIC  & PosMIC & PosTIC & NegMIC & NegTIC \\ \hline
% \multicolumn{1}{c}{\checkmark} & \multicolumn{1}{c|}{\checkmark} & 86.6 & 74.5 & 54.1   & 44.0   & 40.2   & 31.6   \\
%       \multicolumn{1}{c}{\textbf{-}}              & \multicolumn{1}{c|}{\checkmark} & 30.6 & 25.9 & 23.6   & 17.2   & 19.2   & 11.6   \\
% \multicolumn{1}{c}{\checkmark} &  \multicolumn{1}{c|}{\textbf{-}}                              & 84.2 & 72.3 & 52.6   & 42.1   & 46.3   & 37.9   \\ \hline
% \end{tabular}
% \end{adjustbox}
% \end{table}

\subsection{Importance of the \emph{do-operation}}
To prove the importance of the different \emph{do-operation} modules used in our method, we evaluate our model by removing different \emph{do-operation} modules in the architecture. As shown in  \Cref{table:ablation1}, by removing Do-Cause, the model loses the ability of finding causal relationship. Removing Do-effect will lead to performance decrease on NegMIC and NegTIC.

Therefore, the performance on MIC, TIC, PosMIC and PosTIC degrades similar to unsupervised CausalVAE. 
By removing Do-effect and keeping Do-Cause, 
the performance on MIC, TIC, PosMIC and PosTIC significantly improves, while the performance on NegMIC and NegTIC is worse than full model where both cause and effect \emph{do-operation} modules are used.

\begin{table}[!htp]
\centering
\caption{Causal representation metrics of model with different \emph{do-operation} module applied}
\label{table:ablation1}
\renewcommand{\arraystretch}{1.2}
\begin{adjustbox}{width=0.9\textwidth}

\begin{tabular}{cc|cccccccc}
\hlineB{3}
Do-Cause & Do-Effect & MIC $\uparrow$          & TIC $\uparrow$           & PosMIC $\uparrow$         & PosTIC  $\uparrow$       & NegMIC $\downarrow$        & NegTIC $\downarrow$        & \multicolumn{1}{l}{$F_1^{MIC}$ $\uparrow$ } & \multicolumn{1}{l}{$F_1^{TIC}$ $\uparrow$ } \\ \hline
-                   & \checkmark           & 30.6          & 25.9          & 23.6          & 17.2          & \textbf{19.2} & \textbf{11.6} & 36.5                           & 28.8                           \\
\checkmark           & -                   & 84.2          & 72.3          & 52.6          & 42.1          & 46.3          & 37.9          & 53.1                           & 50.2                           \\
\checkmark           & \checkmark           & \textbf{86.6} & \textbf{74.5} & \textbf{54.1} & \textbf{44.0} & 40.2          & 31.6          & \textbf{56.8}                  & \textbf{53.6}                 \\ \hlineB{3}
\end{tabular}

\end{adjustbox}
\end{table}

\subsection{GAE comparison with NOTEARS}
As we mentioned in Section 3, our method incorporate a graph autoencoder (GAE) [17] as causal discovery layer. GAE can learn nonlinear structural causal relationships thus generalizing over NOTEARS [29] which can only learn linear mapping. As shown in \Cref{table:gae}, if we replace GAE with NOTEARS for causal discovery layer, the performance of our model will be harmed since the causal relationships between latent factors can be nonlinear in many cases.

\begin{table}[htp]
\centering
\caption{Causal representation metrics tested on Pendulum and Flow. Higher MIC, TIC, PosMIC and PosTIC value mean better performance. Lower NegMIC and NegTIC value mean better performance. Our methods are trained using only $10\%$ of label.}
\label{table:gae}
\renewcommand{\arraystretch}{1.3}
\begin{adjustbox}{width=\textwidth}
\begin{tabular}{c|cccccccc}
\hlineB{3}
\multirow{2}{*}{Models} & \multicolumn{8}{c}{Pendulum}                  \\ \cline{2-9} 
    & MIC  $\uparrow$         & TIC $\uparrow$           & PosMIC $\uparrow$         & PosTIC $\uparrow$        & NegMIC $\downarrow$        & NegTIC $\downarrow$       & $F_1^{MIC}$ $\uparrow$         & $F_1^{TIC}$ $\uparrow$        \\ \hline
NOTEARS                 & 40.3          & 30.9          & 27.3          & 17.3          & \textbf{26.2} & \textbf{16.2} & 39.6          & 28.7          \\
Our method              & \textbf{86.6} & \textbf{74.5} & \textbf{54.1} & \textbf{44.0} & 40.2          & 31.6          & \textbf{56.8} & \textbf{53.6} \\ \hlineB{3}
\end{tabular}
\end{adjustbox}
\end{table}

\subsection{Synthetic datasets experiments of using a few labels}
In Section 4, we demonstrate our method can outperform other methods which do not use the label, and our method can achieve comparable performance compared with CausalVAE evaluated by PosMIC, PosTIC, NegMIC, and NegTIC. To test our method more comprehensively on synthetic datasets, we conduct experiments of our method, CausalVAE and ConditionVAE using only $10\%$ of labels. As shown in \Cref{table:semi-synthetic}, trained under only $10\%$ of labeled data, CausalVAE and ConditionVAE are difficult to learn either good semantic meaning latent factors which is reflected by MIC and TIC, or attain true causal relationship between cause factors and effect factors, which is shown by PosMIC and PosTIC. As we discussed in Section 4, since CausalVAE fails to encode useful enough semantic factors information, it achieves a low value on NegMIC and NegTIC. ConditionVAE achieves low NegMIC and NegTIC because it aims at learning disentangled latent representation, where each latent factor is enforced to be independent of each other. Thus no causal relationship, correct or wrong, will be learned.

\begin{table}[]
\centering
\caption{Causal representation metrics tested on Pendulum and Flow. Higher MIC, TIC, PosMIC and PosTIC value mean better performance. Lower NegMIC and NegTIC value mean better performance. Our methods are trained using only $10\%$ of label.}
\label{table:semi-synthetic}
\renewcommand{\arraystretch}{1.3}
\begin{adjustbox}{width=\textwidth}
\begin{tabular}{cccccccllccccccll}
\hlineB{3}
\multicolumn{1}{c|}{\multirow{2}{*}{Models}} & \multicolumn{8}{c}{Pendulum}                                                                                                                       & \multicolumn{8}{c}{Flow}                                                                                                      \\ \cline{2-17}
\multicolumn{1}{c|}{}                        & MIC           & TIC           & PosMIC        & PosTIC        & NegMIC        & NegTIC        & $F_1^{MIC}$       & \multicolumn{1}{l|}{$F_1^{TIC}$}        & MIC           & TIC           & PosMIC        & PosTIC        & NegMIC        & NegTIC        & $F_1^{MIC}$        & $F_1^{TIC}$       \\ \hline
\multicolumn{17}{c}{All labels used}                                                                                                                                                                                                                                                                                              \\ \hline
\multicolumn{1}{c|}{CausalVAE [26] }               & \textbf{95.1} & \textbf{81.6} & 53.0          & 43.4          & 46.6          & 37.0          & 53.2          & \multicolumn{1}{l|}{51.4}          & 72.1          & 56.4          & 45.1          & 36.7          & 43.3          & 33.7          & 47.3          & 33.6          \\
\multicolumn{1}{c|}{ConditionVAE [22]}            & 93.8          & 79.6          & 36.5          & 27.8          & 34.6          & 25.7          & 46.9          & \multicolumn{1}{l|}{40.5}          & 75.5          & \textbf{56.5} & 28.6          & 21.3          & 27.2          & 20.6          & 41.1          & 33.6          \\ \hline
\multicolumn{17}{c}{10\% labels used}                                                                                                                                                                                                                                                                                             \\ \hline
\multicolumn{1}{c|}{CausalVAE [26]}               & 64.7          & 55.9          & 39.4          & 30.7          & 37.6          & 28.2          & 48.3          & \multicolumn{1}{l|}{43.0}          & 53.2          & 46.7          & 30.6          & 22.5          & 30.3          & 21.7          & 42.5          & 35.0          \\
\multicolumn{1}{c|}{ConditionVAE [22]}            & 63.2          & 52.1          & 30.5          & 21.3          & \textbf{29.4} & \textbf{24.6} & 42.6          & \multicolumn{1}{l|}{33.2}          & 55.7          & 48.1          & 29.6          & 20.8          & \textbf{26.7} & \textbf{20.1} & 42.1          & 33.0          \\
\multicolumn{1}{c|}{Our method}              & 94.6          & 80.7          & \textbf{70.2} & \textbf{59.5} & 41.2          & 30.4          & \textbf{63.9} & \multicolumn{1}{l|}{\textbf{63.9}} & \textbf{75.7} & 56.1          & \textbf{60.3} & \textbf{51.8} & 37.8          & 29.6          & \textbf{61.2} & \textbf{59.7} \\ \hlineB{3}
\end{tabular}
\end{adjustbox}
\end{table}

\subsection{Experiments detail}

\begin{figure*}[htp]
\centering
\includegraphics[width=\textwidth]{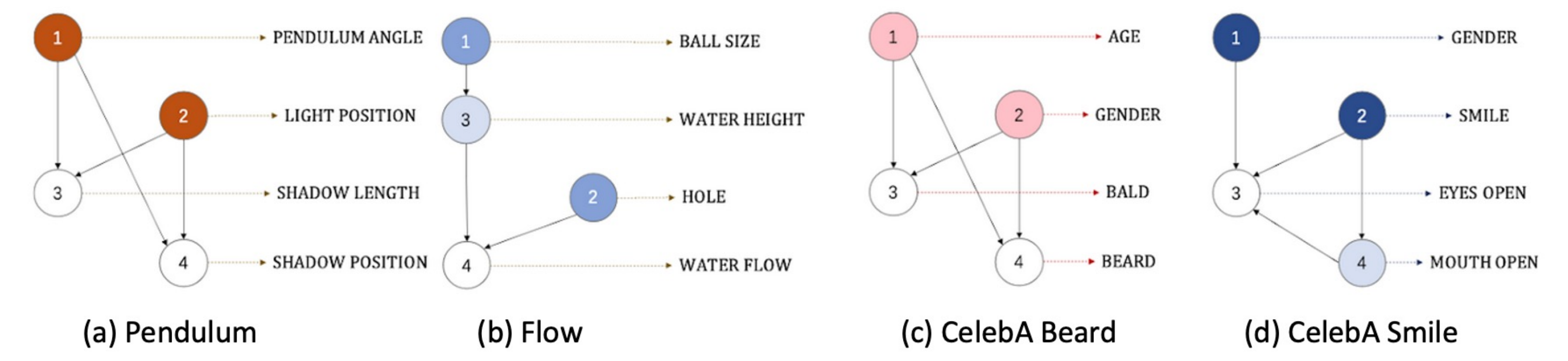}
\caption{Ground truth Causal graph of four datasets.}
\label{fig:true-graph}
\end{figure*}

The true causal graph of each datasets are shown in \Cref{fig:true-graph}.

We use one NVIDIA 1080 Ti GPU as our training and inference device.
Following CausalVAE [26] architecture, we show the VAE architecture of synthetic datasets in \Cref{table:synthetic-vae} and VAE architecture of CelebA dataset in \Cref{table:real-vae}. For latent representation, we also follow the setting of CausalVAE where latent space $z$ is extanded to matrix $z \in R^{n\times k}$ and $n$ is the number of concept and $k$ is latent dimension of each concept.  $k$ is set to $4$ for VAE used in synthetic datasets and $k$ is set to $32$ for VAE used in CelebA dataset. 

As described in Section 3, our loss function for no label training is shown in Equation 10 and the loss for label training is shown in Equation 11. The hyperparameters $(\alpha, \beta, \gamma)$ are grid search among $\{1e^{-3}, 1e^{-2}, 1e^{-1}, 1.0\}$. For training with label, the hyperparameter of $l_u$ is always set to $1$.
\begin{table}[!htp]
\centering
\caption{Synthetic datasets model architecture}
\label{table:synthetic-vae}
\begin{tabular}{clcl}
\hlineB{3}
\multicolumn{2}{c}{encoder}              & \multicolumn{2}{c}{decoder}                      \\ \hline
\multicolumn{2}{c}{4*96*96*900 fc. 1ELU} & \multicolumn{2}{c}{concepts*(4*300 fc. 1ELU)}    \\
\multicolumn{2}{c}{900*300 fc. 1ELU}     & \multicolumn{2}{c}{concepts*(300*300 fc. 1ELU)}  \\
\multicolumn{2}{c}{300*2*concepts*k fc.} & \multicolumn{2}{c}{concepts*(300*1024 fc. 1ELU)} \\
\multicolumn{2}{c}{-}                    & \multicolumn{2}{c}{concepts*(1024*4*96*96 fc.)}  \\ \hlineB{3}
\end{tabular}
\end{table}

\begin{table}[!htp]
\centering
\caption{CelebA datasets model architecture}
\label{table:real-vae}
\begin{tabular}{clcl}
\hlineB{3}
\multicolumn{2}{c}{encoder}                              & \multicolumn{2}{c}{decoder}                                     \\ \hline
\multicolumn{2}{c}{-}                                    & \multicolumn{2}{c}{1*1 conv. 128 1LReLU(0.2), stride 1}         \\
\multicolumn{2}{c}{4*4 conv. 32 1LReLU (0.2), stride 2}  & \multicolumn{2}{c}{4*4 convtranspose. 64 1LReLU(0.2), stride 1} \\
\multicolumn{2}{c}{4*4 conv. 64 1LReLU (0.2), stride 2}  & \multicolumn{2}{c}{4*4 convtranspose. 64 1LReLU(0.2), stride 1} \\
\multicolumn{2}{c}{4*4 conv. 64 1LReLU (0.2), stride 2}  & \multicolumn{2}{c}{4*4 convtranspose. 32 1LReLU(0.2), stride 1} \\
\multicolumn{2}{c}{4*4 conv. 64 1LReLU (0.2), stride 2}  & \multicolumn{2}{l}{4*4 convtranspose. 32 1LReLU(0.2), stride 1} \\
\multicolumn{2}{c}{4*4 conv. 256 1LReLU (0.2), stride 2} & \multicolumn{2}{l}{4*4 convtranspose. 32 1LReLU(0.2), stride 1} \\
\multicolumn{2}{c}{1*1 conv. 3, stride1}                 & \multicolumn{2}{c}{4*4 convtranpose. 3, stride 2}               \\ \hlineB{3}
\end{tabular}
\end{table}

\subsection{Do-operation implementation detail}
\begin{figure}[!htp]
\centering
\includegraphics[width=\textwidth]{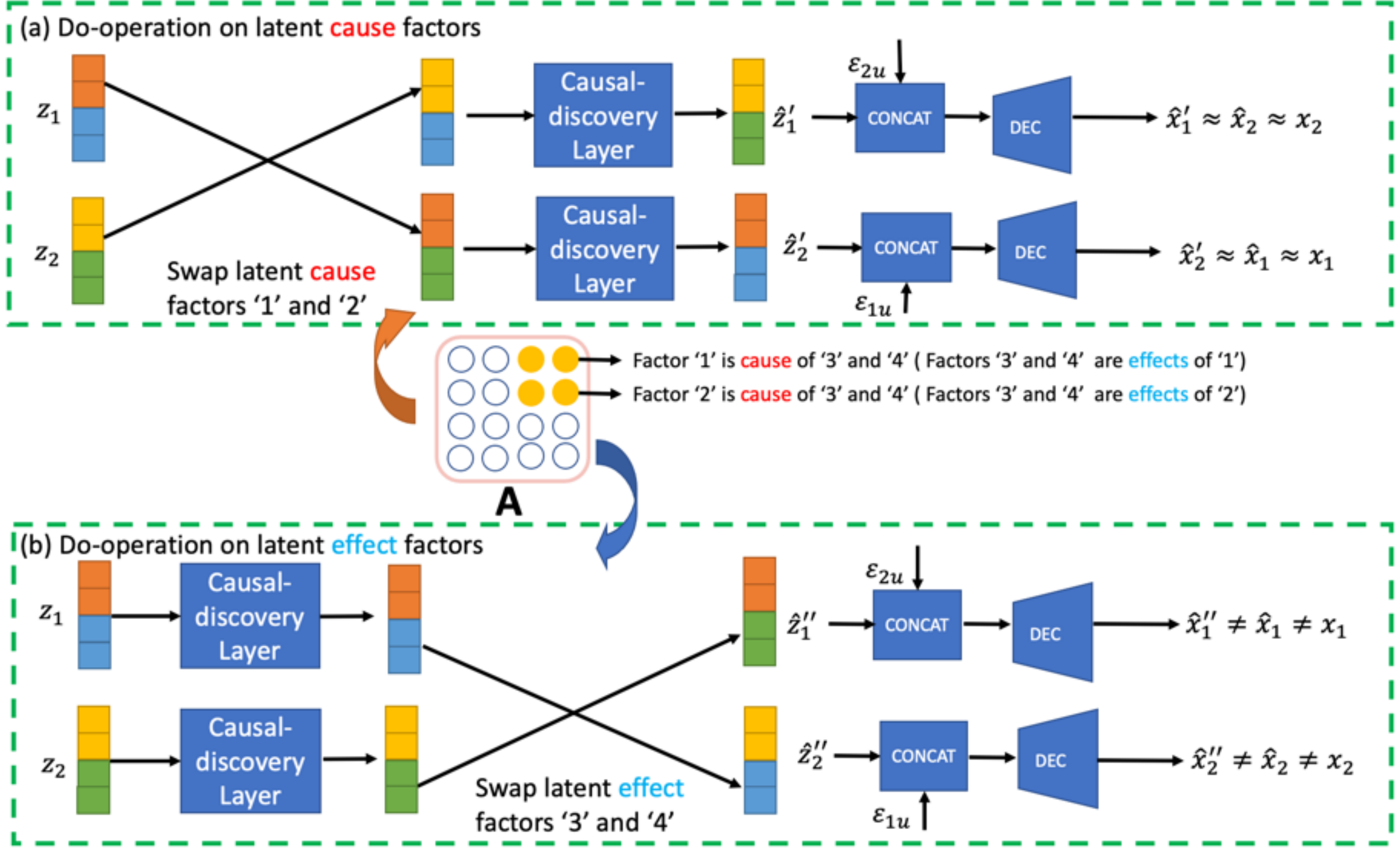}
\caption{\emph{Do-operation} is applied to both cause factors and effect factors. \emph{Do-operation} on cause factors encourage model to learn correct causal relationships and \emph{do-operation} on effect factors prevent model learning wrong causal relationships. }
\label{fig:do-operation-detail}
\end{figure}

As we described in section 3, we apply \emph{do-operation} to both latent cause and effect factors. To better show the implementation of \emph{do-operation} in our work, we describe the process in \Cref{fig:do-operation-detail}.
 As illustrated in \Cref{fig:do-operation-detail}, the cause and effect factors in the latent space are decided by learned causal matrix $A$ which is identical to causal matrix used in causal discovery layer.  After deciding the cause and effect factors, we separately apply \emph{do-operation} on cause and effect factors. Applying \emph{do-operation} to cause factors is straightforward since cause factors have no parent factors and the causal graph stays unchanged. Oppositely, applying \emph{do-operation} to effect factors will both fix the value of effect factors and remove affects from cause factors. Thus, if we swap the effect factors before causal discovery layer, the original causal relationships from cause factors to effect factors still hold. To eliminate the original causal relationships, the swapping operation on effect factors should be applied after causal discovery layer.   
 
 According to [18], \emph{do-operation} replace factors with constants and remove all causal relationships towards the factors. If the label information is available, the \emph{do-operation} is straightforward since the latent factors value can be easily fixed with the label value. However, if the label information is missing, even though the latent factor value can be replaced by some random values, such random values do not guarantee to be meaningful. To obtain the proper constants which replace latent factors, another sample is needed since the reconstruction task force the latent representation encoded from the input are meaningful and can be used as source for \emph{do-operation}.

\subsection{Counter example to prove the weakness of MIC and TIC}
Assuming we have four independent gaussian variables $A$, $B$, $C$ and $D$, where $A\sim \mathcal{N}(\mu_a, \sigma_a^{2})$, $B\sim \mathcal{N}(\mu_b, \sigma_b^{2})$, $C\sim \mathcal{N}(\mu_c, \sigma_c^{2})$ and $D\sim \mathcal{N}(\mu_d, \sigma_d^{2})$. We can create other four gaussian variables $A'$, $B'$, $C'$ and $D'$ where $A'\sim \mathcal{N}(\mu_a, \sigma_a^{2})$,  $B' = \frac{\mu_b}{\mu_a} \cdot A' + ( \sigma_b -\frac{\mu_b}{\mu_a}\sigma_a) \cdot \mathcal{N}(0, 1) $, $C' = \frac{\mu_c}{\mu_a} \cdot A' + ( \sigma_c -\frac{\mu_c}{\mu_a}\sigma_a) \cdot \mathcal{N}(0, 1) $ and $D' = \frac{\mu_d}{\mu_a} \cdot A' + ( \sigma_d -\frac{\mu_d}{\mu_a}\sigma_a) \cdot \mathcal{N}(0, 1) $. By creating new variables like this, it is easy to see that $A'$ has same distribution with $A$, $B'$ has same distribution with $B$, $C'$ has the same distribution with $C$ and $D'$ has the same distribution with $D$. 
Since MIC and TIC only evaluate the marginal distribution of each variable separately, they can not distinguish $A$ from $A'$, $B$ from $B'$, $C$ from $C'$ and $D$ from $D'$. However, $(A,B,C,D)$ have totally different joint distribution from $(A', B', C', D')$. 

\subsection{Metrics implement details}
The superiorities of the proposed new metrics and a simple example has been discussed in Section 4. More details about those new metrics will be discussed in this section. For fully supervised learning or semi-supervised learning method, the metrics calculation are straightforward since every latent elements is controlled by their corresponding label information [5]. For unsupervised methods and our reduced supervision method without using label, we have to first determine the correspondence between each latent factor and each label. We use MIC to choose which latent element represent the label information. As we described in Section 4, MIC can be used to measure the information relevance between a latent factor and a generative label. For each generative factor label, we choose the latent element which achieve maximum MIC value evaluated with that generative factor. After choosing the correspondence between each latent factor with all generative factors label, we can apply Pos/Neg metrics according to the true causal graph provided by the datasets.

\subsection{Reconstruction results}
\begin{figure}[!htp]
    \centering
    \includegraphics[width=\textwidth]{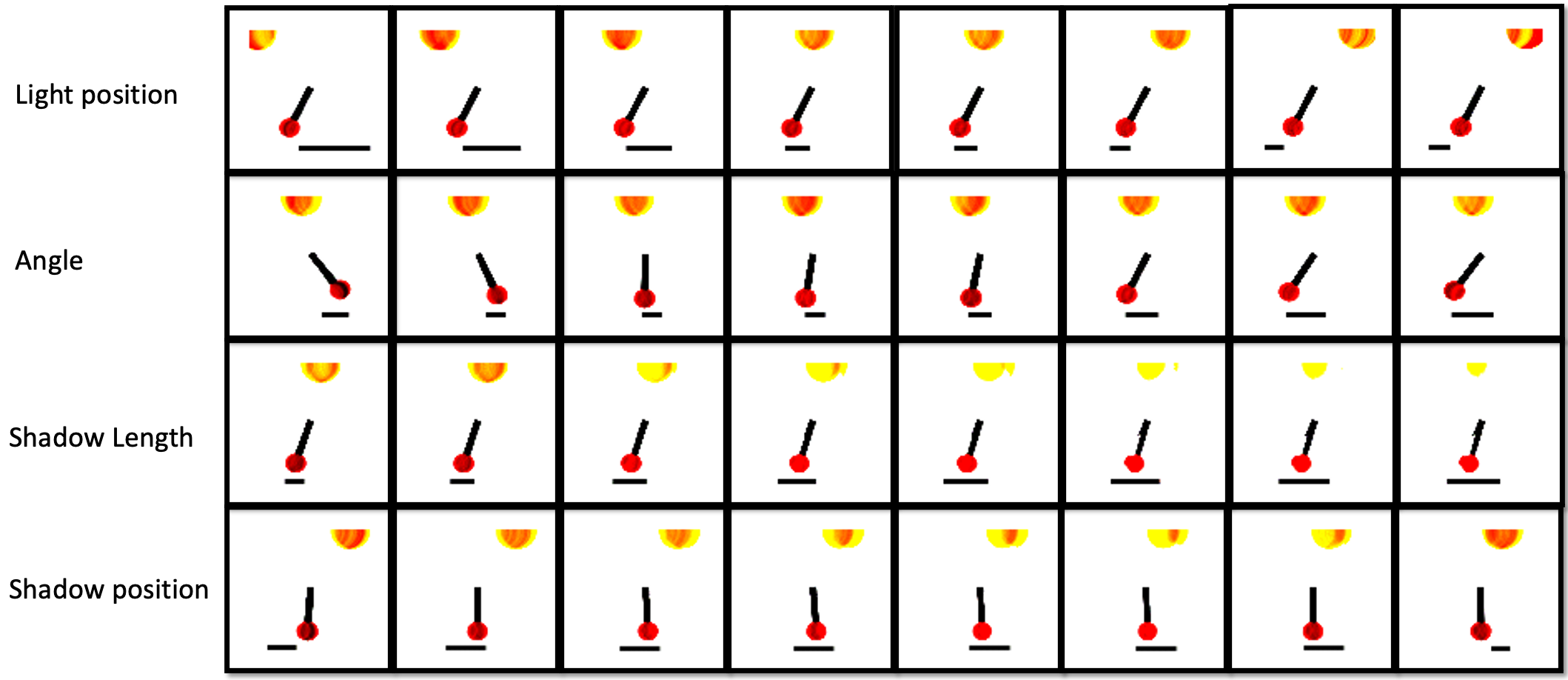}
    \caption{Traversal reconstruction of pendulum dataset. For each rows, we only change one latent factor value and fix all other latent factors. By changing cause factor (light position or angle), we observe corresponding change in effect factors (shadow position and shadow length). Oppositely, by changing effect factor (shadow location and shadow length), the reconstructions can become counterfactual images and the cause factors (light position and angle) stay unchanged.}
    \label{fig:traversal-pendulum}
\end{figure}

\begin{figure}[!htp]
    \centering
    \includegraphics[width=\textwidth]{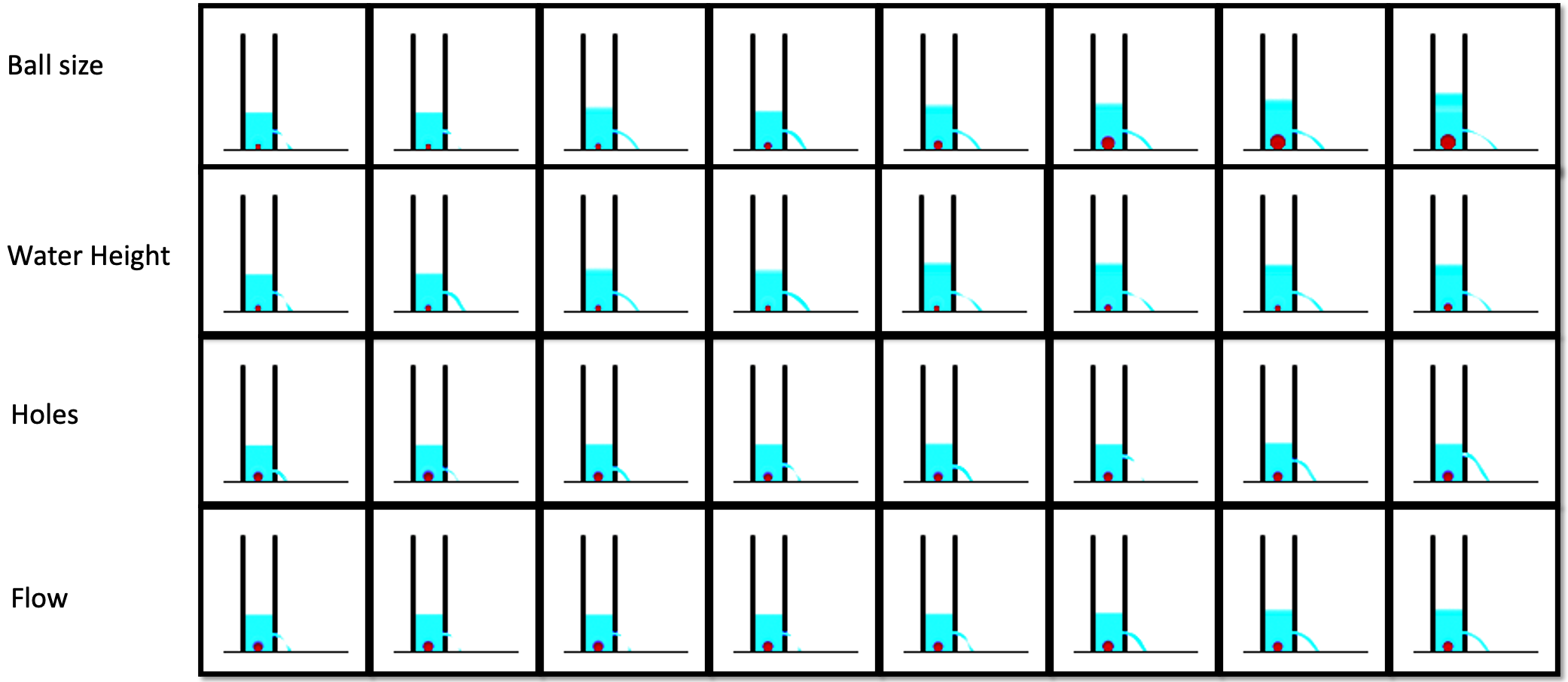}
    \caption{Traversal reconstruction of flow dataset. For each rows, we only change one latent factor value and fix all other latent factors. By changing cause factor (ball size or hole), we observe corresponding change in effect factors (water height and flow). Oppositely, by changing effect factor (water height or flow), the reconstructions can become counterfactual images and the cause factors (ball size and hole) stay unchanged.}
    \label{fig:traversal-flow}
\end{figure}

\begin{figure}[!htp]
    \centering
    \includegraphics[width=0.94\textwidth]{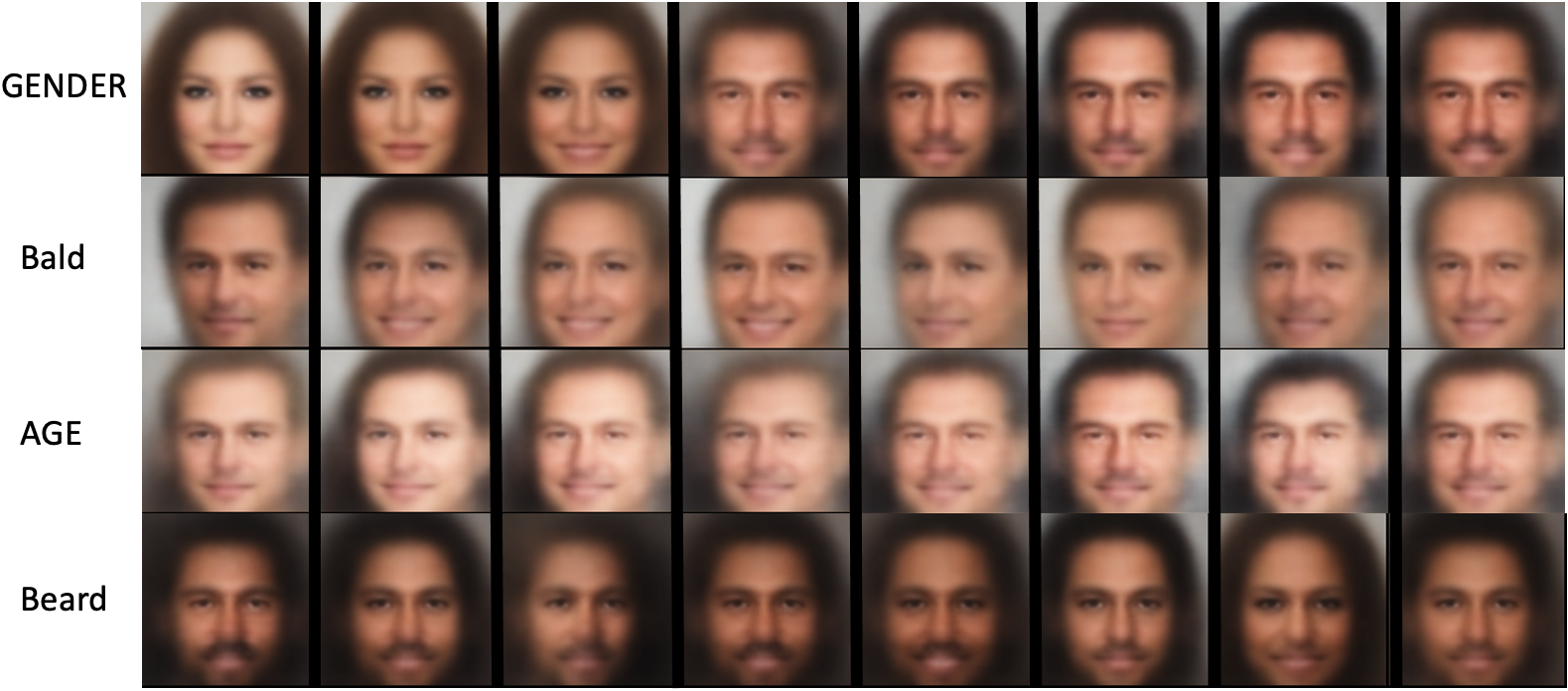}
    \caption{Traversal reconstruction of CelebA(Beard) dataset. For each rows, we only change one latent factor value and fix all other latent factors. By changing cause factor (age or gender), we observe corresponding change in effect factors (bald and beard). Oppositely, by changing effect factor (beard and bald), the reconstructions can become counterfactual images and the cause factors stay unchanged.}
    \label{fig:traversal-celebA-beard}
\end{figure}

\begin{figure}[!htp]
    \centering
    \includegraphics[width=\textwidth]{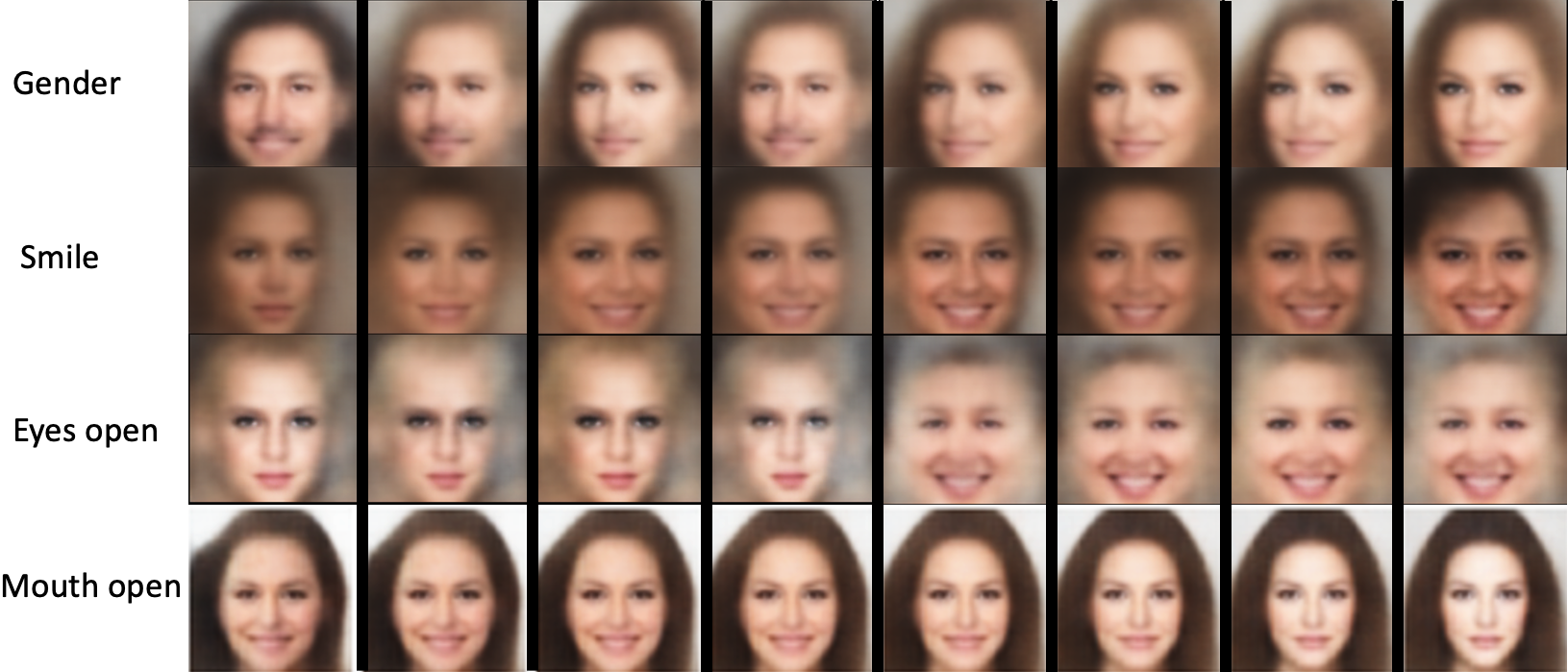}
    \caption{Traversal reconstruction of pendulum CelebA(Smile) dataset. For each rows, we only change one latent factor value and fix all other latent factors. By changing cause factor (gender and smile), we observe corresponding change in effect factors (eyes open). Oppositely, by changing effect factor (shadow eyes open), the reconstructions can become counterfactual images and the cause factors stay unchanged.}
    \label{fig:traversal-celebA-smile}
\end{figure}

We include the image reconstruction results in this section. Shown in \crefrange{fig:traversal-pendulum}{fig:traversal-celebA-smile}, when changing the cause factors, the effect factors shown in reconstructions are changed corresponding. On the contrary, when changing the effect factors, the reconstructions can be counterfactual images and the cause factors stay unchanged.

\end{document}